\documentclass{article}

  \usepackage[preprint]{neurips_2026}

\usepackage[utf8]{inputenc} 
\usepackage[T1]{fontenc}    
\usepackage{hyperref}       
\usepackage{url}            
\usepackage{booktabs}       
\usepackage{amsfonts}       
\usepackage{nicefrac}       
\usepackage{microtype}      
\usepackage{xcolor}         
\usepackage{bbm}
\title{Keep What Audio Cannot Say: \\ Context-Preserving Token Pruning for Omni-LLMs}

\author{%
  Chaeyoung Jung\thanks{Equal contribution}\quad
  Kyeongha Rho\footnotemark[1]\quad
  Joon Son Chung\\
  Korea Advanced Institute of Science and Technology (KAIST)\\
}
\usepackage{amsmath}
\usepackage{booktabs}
\usepackage{amssymb}
\usepackage{multirow}
\usepackage{booktabs}
\usepackage{graphicx}
\usepackage{booktabs}
\usepackage{graphicx}
\usepackage[table]{xcolor}
\usepackage[table]{xcolor}
\usepackage{wrapfig}
\usepackage{caption}
\usepackage{graphicx} 

\usepackage{subcaption} 

\begin{document}

\maketitle

\begin{abstract}
Omnimodal Large Language Models (Omni-LLMs) incur substantial computational overhead due to the large number of multimodal input tokens they process, making token reduction essential for real-world deployment.
Existing Omni-LLM pruning methods typically reduce this cost by selecting tokens that are important for the current query or strongly aligned with cross-modal cues.
However, such strategies can discard evidence that falls outside these criteria, even when needed for different questions or for understanding context beyond aligned audio-visual cues.
To address this limitation, we reframe Omni-LLM token reduction as preserving broad audio-visual context while removing cross-modal redundancy.
We propose \textbf{ContextGuard}, an inference-time token pruning framework built on this principle.
ContextGuard predicts coarse visual semantics from audio and prunes video tokens whose coarse semantics are likely recoverable from audio, while retaining additional video tokens to preserve localized visual details that audio alone cannot specify.
For further compression, our method merges temporally similar video tokens.
The framework requires no downstream LLM fine-tuning and uses only an independently trained lightweight predictor.
On Qwen2.5-Omni and Video-SALMONN2+ at 3B and 7B scales across six audio-visual benchmarks, ContextGuard outperforms prior inference-time pruning methods while pruning more tokens.
Notably, on Qwen2.5-Omni 7B, ContextGuard achieves full-token-level performance on five of six benchmarks while pruning 55\% of input tokens.


\end{abstract}

\section{Introduction}
\label{intro}
Omnimodal large language models (Omni-LLMs)~\cite{chen2023vast, cheng2024videollama, chowdhury2024meerkat, han2024onellm, lyu2023macaw, panagopoulou2023x, ye2024cat, zhan2024anygpt, zhang2023video, zhao2023chatbridge} have rapidly advanced multimodal understanding by jointly processing text, visual, and audio inputs. However, this capability comes with substantial computational overhead, since even short clips can produce thousands of tokens, leading to high memory usage and slow inference. 
While token compression has been widely studied in Video-LLMs~\cite{bai2025qwen25vltechnicalreport,chen2024internvl,jiang2025stormtokenefficientlongvideo,lillava,li2024videochat,lin-etal-2024-video,liu2023llava,qi2025lmm,ye2024fitprunefasttrainingfree,zhang2023video,zhang2024pmod}, typically by removing spatio-temporally redundant tokens or selecting task-relevant tokens, directly transferring this perspective to Omni-LLMs is insufficient.
Unlike Video-LLMs, Omni-LLMs interpret video jointly with audio, and the audio stream often provides compact semantic cues about the same input. As a result, effective Omni-LLM pruning should account for the cross-modal relationship between audio and video, rather than treating video tokens in isolation.

Existing Omni-LLM token pruning methods mainly follow two directions.
\textit{Training-based} approaches~\cite{ding2026omnisift, gong2025echoingpixels} learn compression modules with downstream supervision, sometimes jointly optimizing them with the LLM decoder for task-specific token selection.
In contrast, \textit{training-free} methods such as OmniZip~\cite{zhao2026omnizip} perform inference-time compression by retaining video tokens with high similarity to salient audio tokens, thereby favoring audio-anchored cross-modal evidence.
Despite their differences, these approaches largely cast pruning as selecting an informative subset of tokens, either for a specific task or based on audio-visual (AV) alignment.
These approaches lead to two limitations.
First, training-based selection can become tailored to a specific query or downstream task, limiting the context preserved for different queries.
Second, alignment-based selection can discard evidence outside strongly aligned regions, such as scene text or background objects.
We therefore shift from selecting what appears important to removing redundancy while preserving AV context, including both modality-specific and cross-modal information.

Our approach, \textbf{ContextGuard}, is motivated by the structural asymmetry between audio and video.
A typical Omni-LLM processes hundreds of video tokens per second, but far fewer audio tokens, sometimes as few as two per second in Video-SALMONN2+~\cite{tang2025video}.
This asymmetry reflects their different roles: video carries rich spatio-temporal detail, whereas audio often provides compact semantic cues and speech signals. 
Since audio and video from the same clip can share partial semantics, part of what one modality conveys may be inferred from the other and thus be redundant. While either direction could reduce this redundancy, we use audio to prune video because video accounts for the vast majority of input tokens, and pruning video therefore yields the largest reduction.

Specifically, ContextGuard performs audio-guided video token selection before the downstream LLM decoder. A lightweight audio-to-video semantic predictor (A2V predictor) first estimates coarse visual semantics from audio, and each video token is scored by its similarity to these predicted semantics. ContextGuard prunes tokens whose semantics audio can already convey (e.g., racing cars when the audio contains car racing sounds), while additionally retaining spatially distributed tokens via grid-wise sampling to preserve localized visual attributes that audio alone does not specify (e.g., color and object pose).
For further compression, our method merges temporally similar video tokens. The framework requires no downstream LLM fine-tuning or task-specific pruning supervision. 

\begin{wrapfigure}{r}{0.4\textwidth}
    \vspace{-0.8em}
    \centering
    \includegraphics[width=0.4\textwidth]{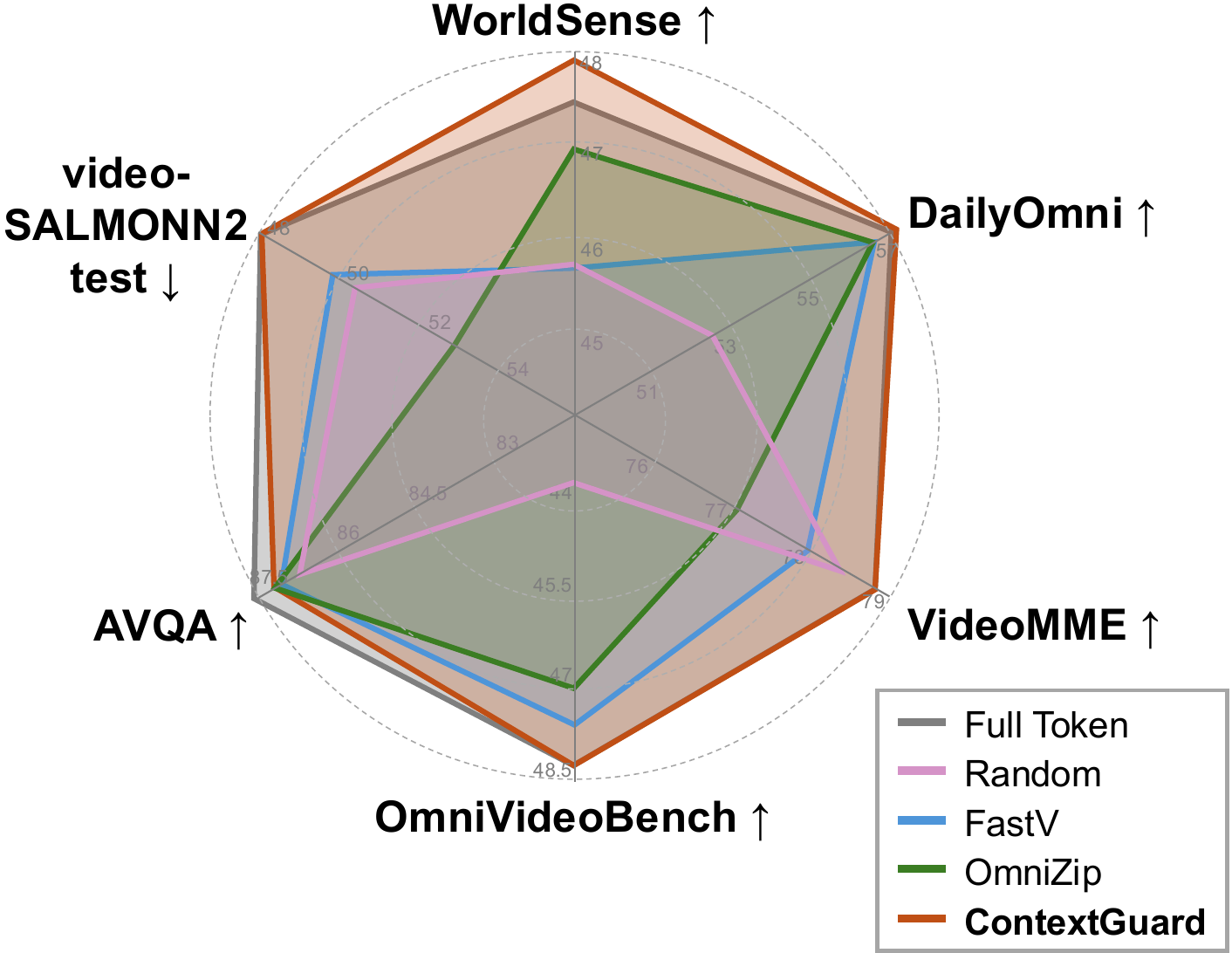}
    \caption{\textbf{Main results on Qwen2.5-Omni 7B.} ContextGuard outperforms previous token compression methods.}
    \label{fig:teaser}
    \vspace{-1em}
\end{wrapfigure}

Experiments on the 3B and 7B variants of Qwen2.5-Omni~\cite{qwen25omni} and Video-SALMONN2+ show that our method outperforms OmniZip~\cite{zhao2026omnizip}, a prior inference-time AV pruning method, in 21 of 24 settings while using fewer input tokens.
On the 7B variant of Qwen2.5-Omni, ContextGuard achieves full-token-level performance on five of six benchmarks while reducing input tokens by about 55\%, as shown in Figure~\ref{fig:teaser} and Table~\ref{tab:main_results}. 
Moreover, although our main method is designed for an offline setting with access to the full sequence, the same principle suggests a simple online-friendly variant that relies only on local temporal information.
This variant remains competitive at comparable compression ratios, suggesting the potential applicability of the proposed pruning principle beyond the offline setting.
Together, these results show that removing what audio already conveys, rather than selecting what appears important, is an effective pruning principle for Omni-LLM token compression.

Our contributions are threefold. First, we shift the framing of Omni-LLM token reduction from selecting important tokens to removing cross-modal redundancy. Second, we propose ContextGuard, an inference-time token pruning framework that instantiates this principle with a lightweight A2V predictor, requiring no downstream LLM fine-tuning. Third, we demonstrate compression–performance gains across 3B and 7B variants of two Omni-LLM backbones on six AV benchmarks.

\section{Method}
\label{Method}

\begin{figure}[t]
    \centering
    \vspace{-3mm}
        \includegraphics[width=\textwidth]{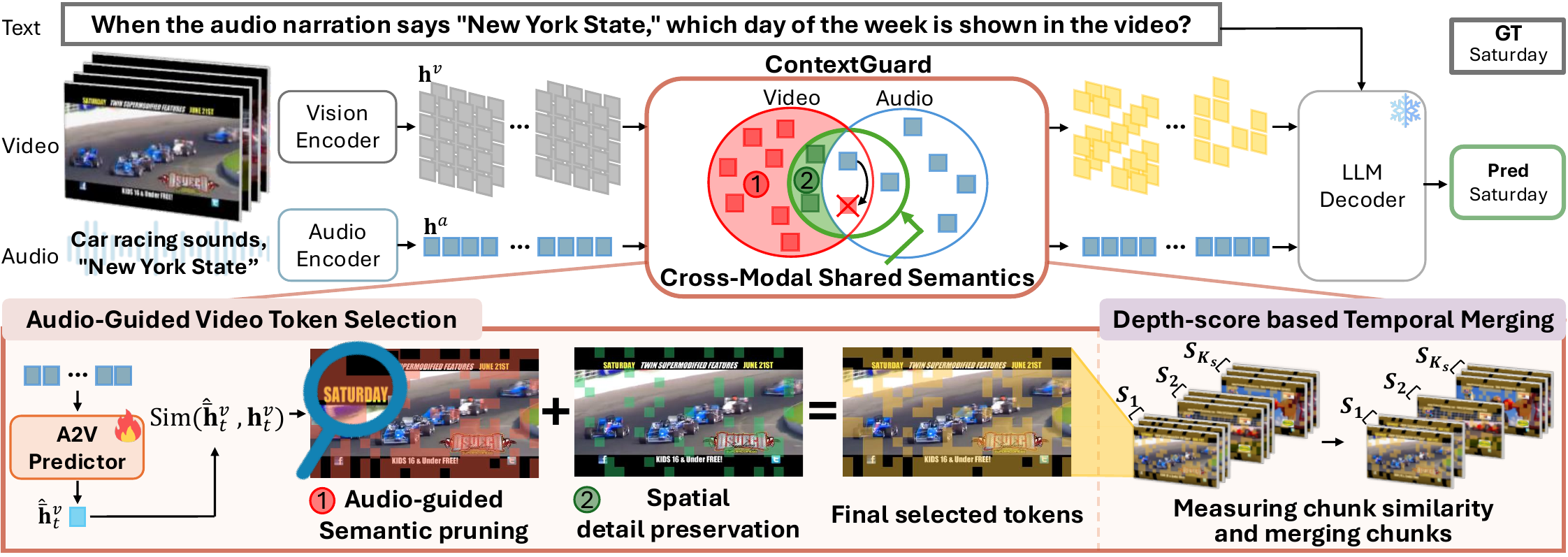}
    \caption{\textbf{Overview of ContextGuard.}
ContextGuard reduces video tokens before the LLM decoder by removing audio-explainable visual redundancy while preserving broad AV context.
For each video chunk $t$ in the interleaved audio-video sequence, an audio-to-video semantic predictor (A2V predictor) estimates coarse visual semantics from the corresponding audio tokens.
ContextGuard performs {audio-guided semantic pruning} by removing tokens predictable from audio (label~\textcircled{\scriptsize 1}) and {spatial detail preservation} by retaining tokens with broad spatial coverage (label~\textcircled{\scriptsize 2}).
The semantic branch preserves visual evidence not explained by audio, such as displayed text, while the spatial branch uses broad spatial coverage to help retain localized visual details.
For further compression, ContextGuard groups similar chunks into temporal segments using depth scores and merges similar chunks within each segment.
Although the audio contains car racing sounds, ContextGuard preserves the non-audio-aligned ``Saturday'' text evidence and answers correctly.}
    \label{fig:main_method}
    \vspace{-4mm}
\end{figure}

\subsection{Motivation and Problem Setup}
To motivate the redundancy-removal perspective from Sec.~\ref{intro}, we view ContextGuard through an information-theoretic lens.
Our goal is to retain visual information that cannot be inferred from the audio alone, while preserving sufficient visual content to represent the original AV context together with the audio.
Let $(V,A)\sim p(V,A)$ denote the full visual and audio token sequences drawn from the data distribution.
Let $\pi$ be a deterministic compression policy that maps $(V,A)$ to a compressed visual sequence $Z_\pi=\pi(V,A)$.
An idealized distribution-level objective is
\begin{equation}
\pi^{*}
=
\operatorname*{arg\,max}_{\pi}
\left\{
\mathcal{I}(V; Z_\pi \mid A)
-
\lambda \mathbb{E}_{(V,A)}[C(Z_\pi)]
\right\},
\label{eq:main_objective}
\end{equation}
where $\mathcal{I}(V; Z_\pi \mid A)$ denotes the conditional mutual information (CMI)~\cite{6773024} between the original visual sequence and the compressed sequence induced by $\pi$, given audio $A$, and $C(Z_\pi)$ denotes the computational cost of a compressed sequence.
We do not optimize Eq.~\eqref{eq:main_objective} directly, since the joint distribution over $(V,A)$ is unknown and the possible pruning and merging decisions are combinatorial.
Instead, we use it as a conceptual guide for what kind of visual information should be retained.

By the definition and symmetry of CMI, the information term admits two equivalent decompositions:
\begin{equation}
\mathcal{I}(V; Z_\pi \mid A)
=
\underbrace{H(V \mid A) - H(V \mid Z_\pi, A)}_{\text{(a) sufficiency view}}
=
\underbrace{H(Z_\pi \mid A) - H(Z_\pi \mid V, A)}_{\text{(b) complementarity view}},
\label{eq:cmi_decomposition}
\end{equation}
where $H(\cdot \mid \cdot)$ denotes conditional entropy.
Form (a) highlights a sufficiency requirement: since $H(V \mid A)$ is fixed with respect to the pruning policy, the compressed sequence $Z_\pi$ should combine with audio $A$ so that $(Z_\pi,A)$ together reduce uncertainty about the original visual sequence $V$.
Form (b) highlights a complementarity requirement.
Since $Z_\pi=\pi(V,A)$ is deterministically obtained from $(V,A)$, we have $H(Z_\pi \mid V,A)=0$, so the information term reduces to $H(Z_\pi \mid A)$.
This suggests that the compressed sequence should contain information not already predictable from audio.
Together, these views suggest retaining visual information not explained by audio while preserving enough content for $Z_\pi$ and $A$ to represent the original visual input.

Motivated by these views, we design a tractable two-stage heuristic that follows this principle without directly optimizing Eq.~\eqref{eq:main_objective}. First, audio-guided video token selection (Sec.~\ref{sec:a2v_token_selection}) uses an audio-to-video semantic predictor to prune video tokens whose semantics are predictable from audio. 
We retain video tokens with low similarity to audio-predicted visual semantics, following the complementarity view of Form (b).
However, audio alone does not specify visual details such as color or object pose, so semantic selection alone can leave such visual information underrepresented. To better satisfy the sufficiency view of Form (a), we add a spatial coverage constraint via grid-wise sampling, which helps preserve such visual details.
Together, the semantic and spatial criteria preserve complementary visual information before temporal compression.
Second, depth-score-based temporal merging (Sec.~\ref{sec:temporal_merge}) reduces the cost term by merging temporally redundant video tokens.

\subsection{Audio-Guided Video Token Selection}
\label{sec:a2v_token_selection}

\paragraph{Audio-guided semantic pruning.}
Let $\mathbf{h}^{v} \in \mathbb{R}^{T \times M \times d}$ and $\mathbf{h}^{a} \in \mathbb{R}^{T \times L \times d}$ denote the visual and audio token features extracted by their encoders.
Here, $T$ is the number of AV chunks, each defined as a contiguous AV block in the interleaved audio-video sequence, e.g., $V^{(t)}$ in $(V^{(1)}, A^{(1)}, V^{(2)}, A^{(2)}, \ldots)$.
Each visual chunk may contain one or a few frames depending on the Omni-LLM tokenizer, and $M$ and $L$ are the numbers of visual and audio tokens per chunk.
We write $\mathbf{h}^{v}_{t,j} \in \mathbb{R}^{d}$ for the $j$-th visual token in chunk $t$.
A lightweight audio-to-video semantic predictor (A2V predictor) $f_{\theta}$, implemented as a learnable-query cross-attention module, takes the audio tokens $\mathbf{h}^{a}_{t,1:L}$ for each chunk and predicts a compact set of coarse visual-semantic embeddings:
\begin{equation}
\hat{\mathbf{h}}^{v}_{t,1:Q}
=
f_{\theta}(\mathbf{h}^{a}_{t,1:L}),
\label{eq:semantic_estimation}
\end{equation}
where $Q$ learnable queries are trained to capture diverse visual-semantic aspects.
We then mean-pool them into a single chunk-level representation $\hat{\bar{\mathbf{h}}}^v_t=\frac{1}{Q}\sum_{q=1}^{Q}\hat{\mathbf{h}}^v_{t,q}$, used as the audio-predicted visual semantics for chunk $t$.
The predictor is trained with contrastive and cosine-similarity-based objectives independently of the downstream LLM and kept frozen during inference. 
Additional implementation details and analyses of the A2V predictor are provided in App.~\ref{supp:a2v_predictor}.

The semantic redundancy of each visual token $\mathbf{h}^{v}_{t,j}$ is estimated by its cosine similarity to the predicted semantics $\hat{\bar{\mathbf{h}}}^{v}_{t}$ for the same chunk:
\begin{equation}
u_{t,j}
=
\mathrm{sim}\!\left(
\mathbf{h}^{v}_{t,j},
\hat{\bar{\mathbf{h}}}^{v}_{t}
\right).
\label{eq:semantic_similarity}
\end{equation}
A larger $u_{t,j}$ indicates that the visual token is more similar to the visual semantics predicted from audio, and is therefore more audio-explainable. We denote the chunk-level score vector as $u_t \in \mathbb{R}^{M}$.
Based on these scores, we retain the $\rho_{\mathrm{sem}}$ fraction of tokens with the lowest similarity scores, marked as label~\textcircled{\scriptsize 1} in Figure~\ref{fig:main_method}, and denote their indices by $\mathcal{P}^{(t)}_{\mathrm{sem}}$.
The retained tokens are more likely to contain visual evidence that is not predictable from audio.
We analyze the effect of $\rho_{\mathrm{sem}}$ in App.~\ref{supp;sem_ratio}. Specifically, we vary $\rho_{\mathrm{sem}}$ and measure the resulting compression rate and first-token KL divergence between the LLM output distributions under full-token and pruned inputs.


\paragraph{Spatial detail preservation.}
While audio-guided semantic pruning retains tokens whose coarse semantics are not predictable from audio, this criterion may miss localized visual details such as color or object pose in audio-explainable regions.
To recover such details that audio alone cannot specify, we add a complementary spatial branch that retains tokens with broad spatial coverage of the visual input (label~\textcircled{\scriptsize 2} in Figure~\ref{fig:main_method}).

For each video chunk with $F$ frames and an $H \times W$ spatial grid, we have $M = FHW$ visual tokens.
We first average token embeddings over frames, yielding a single $H \times W$ feature map per chunk for spatial selection.
We then apply grid-wise sampling, partitioning the grid into local cells and retaining one token per cell to encourage spatial coverage.
Within each cell, we select the token with the largest local spatial variation, computed as the sum of $\ell_2$-norm differences to its horizontal and vertical grid neighbors.
This variation is used only to choose a representative token within each cell, while the overall number of selected tokens is controlled by $\rho_{\mathrm{spa}}$, the spatial retention ratio.

The selected spatial indices are repeated across all frames in the chunk, yielding $\mathcal{P}^{(t)}_{\mathrm{spa}}$.
Additional details and analyses of grid-wise sampling and spatial token selection are provided in App.~\ref{supp:spa_ratio}.
The semantic and spatial selections are combined to yield the final selected token indices for each chunk:
\begin{equation}
\mathcal{P}^{(t)}_{\mathrm{sel}}
=
\mathcal{P}^{(t)}_{\mathrm{sem}}
\cup
\mathcal{P}^{(t)}_{\mathrm{spa}}.
\label{eq:framewise_selected_tokens}
\end{equation}
These indices define the retained video tokens before temporal compression.

\subsection{Depth-Score-Based Temporal Merging}
\label{sec:temporal_merge}
Temporally similar chunks are often redundant and can be further compressed.
However, independently selected tokens across chunks may have different layouts, making direct chunkwise merging unstable.
We therefore group temporally similar chunks into segments and enforce a shared token index set within each segment before merging.
For each chunk $t$, we compute mean-pooled video and audio representations
$\bar{\mathbf{h}}^{v}_{t}=\frac{1}{M}\sum_{j}\mathbf{h}^{v}_{t,j}$ and
$\bar{\mathbf{h}}^{a}_{t}=\frac{1}{L}\sum_{\ell}\mathbf{h}^{a}_{t,\ell}$,
and define adjacent-chunk similarities $s^{m}_{t}=\mathrm{sim}(\bar{\mathbf{h}}^{m}_{t},\bar{\mathbf{h}}^{m}_{t-1})$ for $m \in \{v,a\}$.
Following~\cite{shu2025video}, we compute the depth score
\begin{equation}
d^{m}_t
=
\max_{i < t} s^{m}_i + \max_{i > t} s^{m}_i - 2s^{m}_t,
\label{eq:depth_score}
\end{equation}
which marks valleys in temporal similarity as segment boundary candidates. The union of boundary candidates from both modalities yields segments $\mathcal{S}=\{\mathcal{S}_k\}_{k=1}^{K_s}$.

For each segment $\mathcal{S}_k$, we average $u_t$ across chunks to obtain $\bar{u}^{(k)}$, reflecting chunk-level changes in audio-conditioned visual redundancy.
For efficiency, we reuse the spatial selection $\mathcal{P}^{(t_k)}_{\mathrm{spa}}$ from the first chunk $t_k \in \mathcal{S}_k$, since visual frames within a segment are highly similar.
We then apply semantic selection using $\bar{u}^{(k)}$ and combine it with the reused spatial indices, yielding a shared index set $\mathcal{P}^{(k)}_{\mathrm{sel}}$ for all chunks in the segment.
Within each $\mathcal{S}_k$, neighboring chunks whose visual similarity exceeds $\tau_{\mathrm{merge}}$ are merged by averaging their retained token embeddings.
The resulting tokens form $Z$ and are fed to the LLM decoder in the original interleaved order, preserving positional structure and full audio tokens.
Further details on boundary selection and merging are provided in App.~\ref{supp:depth_score}.


\section{Experiments}
\label{Experiments}
We evaluate ContextGuard from four perspectives. We first compare it with prior inference-time token pruning methods in terms of token compression and downstream task performance (Table~\ref{tab:main_results}). We then verify whether the learned A2V predictor captures coarse visual semantics (Table~\ref{tab:retrieval_predictor}) and analyze the core pruning components of ContextGuard (Tables~\ref{tab:ablation_build_up}--\ref{tab:ablation_depth}). We also report practical inference statistics for the main offline framework and evaluate a simple online-friendly variant (Tables~\ref{tab:efficiency_wrap} and~\ref{tab:online_variant}).

\subsection{Experimental setup}
\paragraph{Benchmarks and evaluation.}
We evaluate on six AV benchmarks: WorldSense~\cite{hong2025worldsense} (World.), a real-world omnimodal task; Daily-Omni~\cite{zhou2025daily} (Daily.), which tests cross-modal temporal reasoning in daily-life videos; Video-MME~\cite{fu2025video}, for general video understanding; AVQA~\cite{yang2022avqa}, for audio-visual QA over sounding objects and interactions; OmniVideoBench~\cite{li2025omnivideobench} (OmniVid.), for synergistic multimodal reasoning with modality complementarity and long-term temporal understanding; and the video-SALMONN2 test set~\cite{tang2025video} (video-SAL2.), for AV captioning. We report accuracy on the first five benchmarks and the official total error rate on video-SAL2. (lower is better).\footnote{\url{https://huggingface.co/datasets/videoSALMONN2/video-SALMONN_2_testset}}
For World., Video-MME, and OmniVid., we evaluate samples shorter than 1 minute, as running the full-token reference on the complete benchmark exceeds our GPU memory budget.

\paragraph{Implementation details.}
We implement ContextGuard on Qwen2.5-Omni~\cite{qwen25omni} (7B/3B) and Video-SALMONN2+~\cite{tang2025video} (7B/3B), and run all experiments on NVIDIA A6000 48GB GPUs.
We follow each backbone's default maximum per-frame pixel setting.
Across all models and benchmarks, we use the same fixed hyperparameters: $\rho_{\mathrm{sem}}=0.5$, $\rho_{\mathrm{spa}}=0.1$, and $\tau_{\mathrm{merge}}=0.98$.
Additional hyperparameter sweep ablations are provided in App.~\ref{supp:hyper_analysis}.

\paragraph{Baselines.}
We compare with four settings: \textit{Full Token}, which uses the original unpruned input; \textit{Random}, which prunes randomly under a fixed compression ratio; \textit{FastV}~\cite{chen2024image}, extended to AV inputs by applying its attention-based pruning criterion to the audio-video token sequence; and \textit{OmniZip}~\cite{zhao2026omnizip}, a recent inference-time AV pruning method that uses audio cues to preserve aligned AV event tokens.

\subsection{Main results}
\begin{table*}[t]
\centering
\footnotesize
\setlength{\tabcolsep}{3.8pt}
\renewcommand{\arraystretch}{0.93}
\caption{\textbf{Main results on four AV-LLMs.} ContextGuard achieves strong token compression while largely preserving full-token performance. Full-token results are shown in gray as reference, and boldface compares pruning methods only. Comp. denotes the average token compression ratio. Avg. is computed by normalizing each benchmark so that the full-token result is 100 and then averaging across all six benchmarks; for video-SAL2., where lower is better, we use the inverse ratio.}
\vspace{-2mm}
\label{tab:main_results}
\begin{tabular}{lcccccccc}
\toprule
Method & Comp.$\uparrow$ (\%) & World.$\uparrow$ & Daily.$\uparrow$ & Video-MME$\uparrow$ & OmniVid.$\uparrow$ & AVQA$\uparrow$ & video-SAL2.~$\downarrow$ & Avg.$\uparrow$ (\%) \\
\midrule

\multicolumn{9}{c}{\textit{Qwen2.5-Omni 7B}} \\
\cmidrule(lr){1-9}
\rowcolor{gray!10}
Full Token  & 0  & 47.4 & 57.1 & 78.8 & 48.2 & 87.6 & 48.1 & 100.0 \\
Random      & 50 & 45.7 & 52.4 & 78.4 & 43.4 & 86.7 & 50.4 & 95.4 \\
FastV       & 50 & 45.6 & 56.6 & 77.9 & 47.6 & 87.0 & 49.8 & 98.1 \\
OmniZip     & 54 & 46.8 & 56.6 & 77.1 & 47.0 & 87.0 & 52.8 & 97.3 \\
\textbf{ContextGuard}        & \textbf{55} & \textbf{47.7} & \textbf{57.2} & \textbf{78.8} & \textbf{48.2} & \textbf{87.1} & \textbf{48.1} & \textbf{100.0} \\
\midrule

\multicolumn{9}{c}{\textit{Qwen2.5-Omni 3B}} \\
\cmidrule(lr){1-9}
\rowcolor{gray!10}
Full Token  & 0  & 47.7 & 57.7 & 75.8 & 44.0 & 87.6 & 53.5 & 100.0 \\
Random      & 50 & 44.1 & 53.1 & 74.0 & 42.8 & 87.1 & 56.1 & 95.7 \\
FastV       & 50 & 46.7 & 55.6 & 74.0 & 44.0 & 87.2 & 54.6 & 98.2 \\
OmniZip     & 54 & 47.1 & 55.8 & 74.9 & 42.8 & \textbf{87.3} & 52.4 & 98.9 \\
\textbf{ContextGuard}        & \textbf{61} & \textbf{47.7} & \textbf{56.6} & \textbf{75.8} & \textbf{45.2} & \textbf{87.3} & \textbf{52.3} & \textbf{100.5} \\
\midrule

\multicolumn{9}{c}{\textit{Video-SALMONN2+ 7B}} \\
\cmidrule(lr){1-9}
\rowcolor{gray!10}
Full Token  & 0  & 50.7 & 56.3 & 79.2 & 43.4 & 67.8 & 18.9 & 100.0 \\
Random      & 50 & 47.3 & 51.6 & 76.2 & 39.5 & 66.8 & 25.7 & 90.7 \\
FastV       & 50 & 46.3 & 51.8 & 79.2 & 42.2 & 66.1 & 24.0 & 92.8 \\
OmniZip     & 49 & 47.8 & 53.8 & 79.2 & 42.8 & \textbf{68.8} & 22.4 & 95.7 \\
\textbf{ContextGuard}        & \textbf{55} & \textbf{50.6} & \textbf{55.5} & \textbf{81.4} & \textbf{47.0} & 66.9 & \textbf{19.9} & \textbf{100.5} \\
\midrule

\multicolumn{9}{c}{\textit{Video-SALMONN2+ 3B}} \\
\cmidrule(lr){1-9}
\rowcolor{gray!10}
Full Token  & 0  & 48.1 & 57.7 & 76.2 & 45.8 & 81.6 & 22.7 & 100.0 \\
Random      & 50 & 44.3 & 51.6 & 73.2 & 42.0 & 81.2 & 30.3 & 90.6 \\
FastV       & 50 & 45.6 & 53.2 & 71.0 & 36.1 & 81.0 & 27.4 & 90.2 \\
OmniZip     & 49 & 45.6 & 54.7 & 73.2 & 41.6 & 81.3 & 26.3 & 93.7 \\
\textbf{ContextGuard}        & \textbf{56} & \textbf{47.1} & \textbf{55.5} & \textbf{73.6} & \textbf{44.6} & \textbf{81.4} & \textbf{24.2} & \textbf{96.9} \\
\bottomrule
\end{tabular}
\vspace{-3mm}
\end{table*}
\paragraph{Quantitative results.}
To evaluate whether our method can aggressively reduce the input token budget while maintaining downstream performance, we apply it to two representative Omni-LLMs, Qwen2.5-Omni and Video-SALMONN2+, each with 7B and 3B variants, and evaluate it on six AV benchmarks. As shown in Table~\ref{tab:main_results}, our method generally outperforms existing baselines. 
We compute average normalized performance by setting each full-token result to 100 and averaging relative scores, using the inverse ratio for video-SAL2.
Notably, our method reaches near full-token normalized performance on Qwen2.5-Omni 7B/3B and Video-SALMONN2+ 7B while pruning more tokens. The advantage is particularly clear on the video-SAL2 test set, where previous pruning methods suffer substantial degradation on the captioning task while our method remains closer to full-token performance. These results support our central claim that AV token pruning can remove audio-redundant visual information while preserving broad  AV context.
App.~\ref{supp:dailyomni_breakdown} further shows that
ContextGuard is especially effective on Daily. categories requiring broader AV context.

\begin{figure}[t]
    \centering
        \centering
        \includegraphics[width=\textwidth]{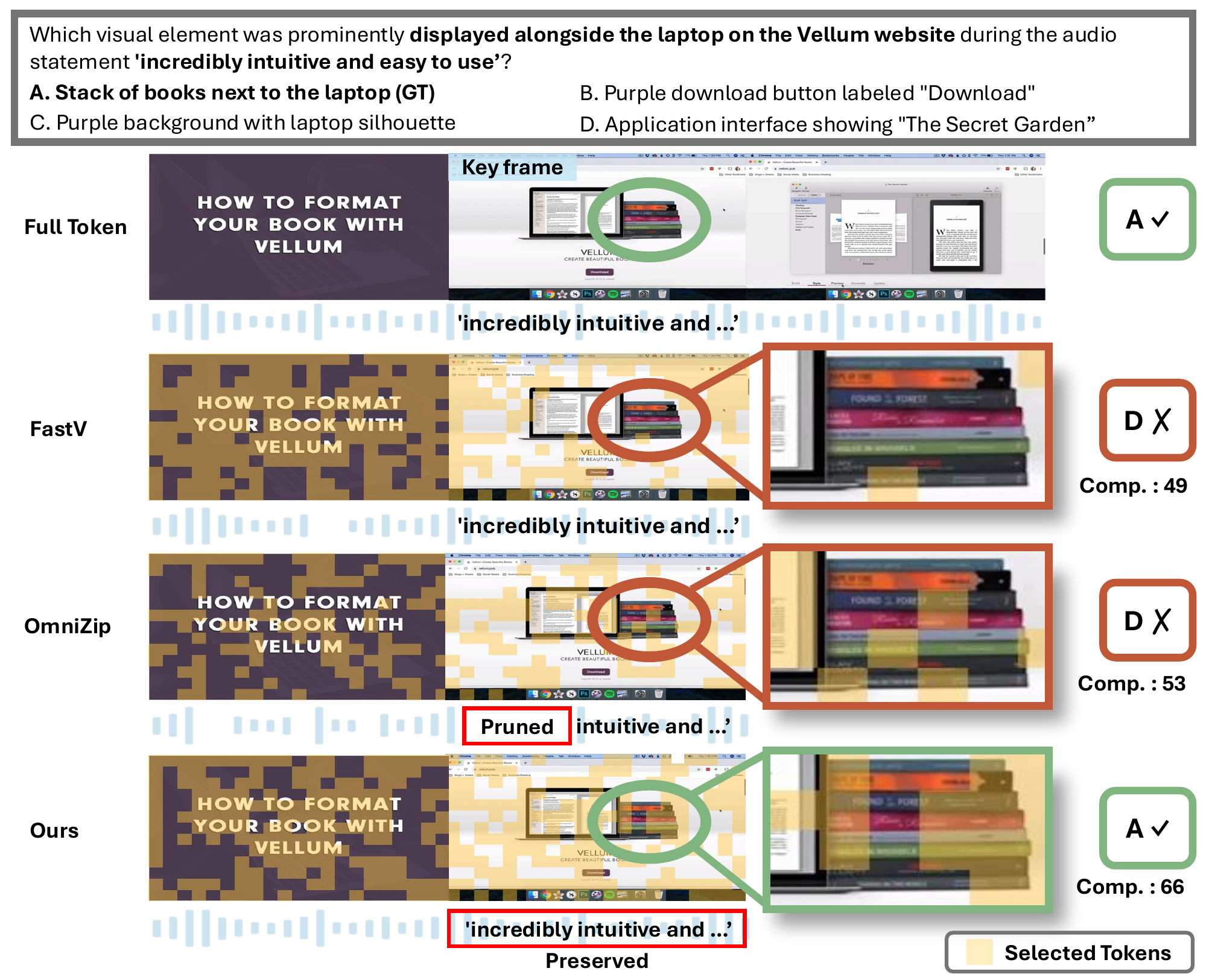}
    \caption{\textbf{Main qualitative results.} FastV and OmniZip fail to preserve visual evidence that is not directly aligned with the audio narration or with the most salient objects in the video, resulting in incomplete context. In contrast, ContextGuard preserves such non-audio-aligned visual information, maintains broad AV context under aggressive token compression, and recovers the correct answer.}
    \label{fig:main_qual1}
  
\end{figure}

\paragraph{Qualitative results.}
Figure~\ref{fig:main_qual1} provides a qualitative example on the Daily. benchmark using Qwen2.5-Omni 7B. Answering the question requires both precise speech localization, specifically identifying the narration segment beginning with ``incredibly intuitive and ...'', and recognition of the visual object shown next to the laptop in the key frame. FastV and OmniZip fail to preserve the stack of books beside the laptop, likely because it is neither strongly audio-aligned nor among the most visually salient objects. Moreover, OmniZip removes the crucial speech segment containing ``incredibly.'' In contrast, our method preserves both the relevant audio cue and the non-audio-aligned visual detail, maintaining the overall context and recovering the correct answer under more aggressive token compression.
Additional qualitative results and failure-case analysis are provided in App.~\ref{supp:downstream_qual}.

\subsection{Analysis of the A2V predictor}

\begin{wraptable}{r}{0.48\textwidth}
\vspace{-1em}
\centering
\footnotesize
\setlength{\tabcolsep}{2.6pt}
\renewcommand{\arraystretch}{0.92}
\caption{\textbf{A2V predictor analysis.}
The learned A2V embeddings improve audio-to-video retrieval and downstream task performance over the original audio embeddings.}
\vspace{-2mm}
\begin{tabular}{lccccc}
\toprule
Method & R@1$\uparrow$ & R@5$\uparrow$ & MedR$\downarrow$ & World. & Daily.\\
\midrule
\multicolumn{6}{c}{\textit{Qwen2.5-Omni 7B}} \\
\cmidrule(lr){1-6}
orig & 5.8 & 17.8 & 44 & 47.6 &56.7  \\
ours & \textbf{12.9} & \textbf{36.9} & \textbf{11} & \textbf{47.7} &\textbf{57.2}  \\
\midrule
\multicolumn{6}{c}{\textit{Video-SALMONN2+ 7B}} \\
\cmidrule(lr){1-6}
orig & 2.3 & 8.4 & 68 & 47.9 & 54.2 \\
ours & \textbf{8.7} & \textbf{24.0} & \textbf{21} &\textbf{50.6}  &\textbf{55.5}  \\
\bottomrule
\end{tabular}
\label{tab:retrieval_predictor}
\vspace{-1em}
\end{wraptable}

To analyze the A2V predictor, we evaluate audio-to-video retrieval on the VGGSound~\cite{chen2020vggsound} test set and compare World. and Daily. performance using either the original audio embeddings (\textit{orig}) or the A2V predictor embeddings (\textit{ours}).
For pruning, \textit{orig} replaces the A2V-predicted visual representation in Eq.~\eqref{eq:semantic_similarity} with the original audio embedding under the same low-similarity selection rule.
As shown in Table~\ref{tab:retrieval_predictor}, the A2V predictor embeddings consistently improve Recall@1/5 and substantially reduce the median rank (MedR), indicating stronger alignment with visual semantics.
The lower World. and Daily. performance of \textit{orig} further supports the predictor embedding as a practical signal for estimating audio-redundant visual semantics.
Additional retrieval details and qualitative examples are provided in App.~\ref{supp:a2v_retrieval}.

\subsection{Ablation studies}
We analyze the core components of ContextGuard through ablations. Unless otherwise noted, all ablation studies are conducted on Qwen2.5-Omni 7B and Video-SALMONN2+ 7B using the World. and Daily. benchmarks. We also report first-token-generation KL divergence between full-token and pruned output distributions as a complementary fidelity measure of the original model behavior.

\begin{table*}[t]
\centering
\footnotesize
\setlength{\tabcolsep}{3.5pt}
\renewcommand{\arraystretch}{0.95}

\begin{minipage}[t]{0.51\textwidth}
\centering
\vspace{-2mm}
\captionof{table}{\textbf{Cumulative component ablation.} We evaluate the cumulative effect of adding semantic pruning, spatial details, and depth-based merging.}
\label{tab:ablation_build_up}
\vspace{-0.8em}
\begin{tabular}{lccc}
\toprule
Variant & World. & Daily. & Comp.   \\
\midrule
\multicolumn{4}{c}{\textit{Qwen2.5-Omni 7B}} \\
\cmidrule(lr){1-4}
\rowcolor{gray!10}
Full Token & 47.4 & 57.1 & -- \\
+ Semantic pruning & 47.7 & 56.4 & 46  \\
+ Spatial details & \textbf{48.1} & 56.8 & 40  \\
+ Depth-based merging & 47.7 & \textbf{57.2} & \textbf{52}  \\
\midrule
\multicolumn{4}{c}{\textit{Video-SALMONN2+ 7B}} \\
\cmidrule(lr){1-4}
\rowcolor{gray!10}
Full Token & {50.7} & 56.3 & -- \\
+ Semantic pruning & 50.4 & 55.2 & 50  \\
+ Spatial details & \textbf{50.8} & {55.3} & 43  \\
+ Depth-based merging & 50.6 & \textbf{55.5} & \textbf{54} \\
\bottomrule
\end{tabular}
\vspace{-2mm}
\end{minipage}
\hfill
\begin{minipage}[t]{0.46\linewidth}
\centering
\setlength{\tabcolsep}{3.2pt}
\renewcommand{\arraystretch}{1.0}
\vspace{-2mm}
\caption{\textbf{Semantic token selection ablation.}
We compare random, high, and low-similarity semantic token selection. Low-similarity selection performs best across backbones.}
\vspace{-0.4em}
\label{tab:low_semantic_tokens}
\begin{tabular}{lcccc}
\toprule
Method & World. & Daily. & Comp.  & KL$\downarrow$ \\
\midrule
\multicolumn{5}{c}{\textit{Qwen2.5-Omni 7B}} \\
\cmidrule(lr){1-5}
Random & 47.4 & 56.4 & 52 & 0.039 \\
High & 45.2 & 53.8 & 51 & 0.079 \\
Low (ours) & \textbf{47.7} & \textbf{57.2} & \textbf{52} & \textbf{0.028} \\
\midrule
\multicolumn{5}{c}{\textit{Video-SALMONN2+ 7B}} \\
\cmidrule(lr){1-5}
Random & 50.3 & 55.2 & 54 & 0.008 \\
High & 49.4 & 54.5 & 54 & 0.013 \\
Low (ours) & \textbf{50.6} & \textbf{55.5} & \textbf{54} & \textbf{0.007} \\
\bottomrule
\end{tabular}
\end{minipage}
\vspace{-2mm}
\end{table*}

\paragraph{Component-wise analysis.}
We first examine how the three main components of our framework, namely audio-guided semantic pruning, spatial detail preservation, and depth-score-based temporal merging (depth-based merging), contribute to downstream task performance and the compression rate. Table~\ref{tab:ablation_build_up} shows that semantic pruning alone already provides strong compression with only a marginal performance drop, suggesting that the coarse semantics of a substantial portion of visual tokens are predictable from audio and therefore redundant. Adding the spatial branch consistently improves performance at the cost of a slightly lower compression ratio, confirming the importance of retaining localized visual details that audio alone does not specify. Finally, depth-based merging pushes compression beyond 50\% with comparable downstream performance, indicating that similarity-based merging removes inter-chunk redundancy while preserving key scene information.

\paragraph{The role of low-semantic-similarity tokens.}
To analyze the role of retaining tokens with low similarity to the audio-predicted visual semantics in preserving broad  AV context, Table~\ref{tab:low_semantic_tokens} compares three variants that differ only in the semantic token selection rule, while keeping spatial detail preservation and depth-based merging unchanged.
\textit{Random} selects semantic tokens randomly under a matched compression ratio, while \textit{High} retains tokens with high similarity to the audio-predicted visual semantics, and \textit{Low} denotes our low-semantic-similarity token selection rule.
Retaining low-semantic-similarity tokens consistently outperforms both random token selection and high-similarity token selection on both Omni-LLMs, while yielding lower KL divergence to the full-token output distribution.
These results support our key insight that many audio-explainable visual tokens contain coarse semantics redundant with the compact audio stream.
Low-similarity tokens instead preserve visual evidence not predicted from audio.
Additional qualitative results showing that low-semantic-similarity tokens capture non-audio-aligned regions are provided in App.~\ref{supp:qual_semantic}.

\paragraph{Ablation of temporal grouping and merging.}
We compare our depth-based merging strategy with two simpler baselines.
\textit{Fixed segmentation} groups every three consecutive chunks into one segment.
\textit{Depth-based pruning} uses the same depth-score-based segmentation as ours, but retains only the first representative chunk from each similar chunk group instead of merging retained tokens.
As shown in Table~\ref{tab:ablation_depth}, both alternatives are less reliable than our method.
Fixed segmentation often causes larger performance drops, suggesting that naive temporal partitions do not align well with semantic changes in the video.
Depth-based pruning avoids token-layout merging, but discards entire chunks.
As a result, it can lose useful cues that remain within similar chunk groups, particularly for Video-SALMONN2+ 7B.
In contrast, our method detects local similarity changes in both audio and video to form temporal segments.
It then enforces a shared retained token layout within each segment, enabling stable temporal merging under aggressive pruning.
This yields the best compression--accuracy trade-off across different Omni-LLMs.

\begin{table}[t]
\centering
\footnotesize

\begin{minipage}[t]{0.48\textwidth}
\centering
\captionof{table}{\textbf{Temporal compression ablation.} We compare depth-based merging with fixed segmentation and depth-based pruning.}
\renewcommand{\arraystretch}{0.95}
\label{tab:ablation_depth}
\begin{tabular}{lccc}
\toprule
Variant & World. & Daily. & Comp. \\
\midrule
\multicolumn{4}{c}{\textit{Qwen2.5-Omni 7B}} \\
\cmidrule(lr){1-4}
\rowcolor{gray!10}
Full Token & 47.4 & 57.1 & -- \\
Fixed segmentation  &47.3 & 56.9 & 51 \\
Depth-based pruning  & \textbf{47.9} & 56.9 & 52 \\
Depth-based merging  & 47.7 & \textbf{57.2} & \textbf{52} \\
\midrule
\multicolumn{4}{c}{\textit{Video-SALMONN2+ 7B}} \\
\cmidrule(lr){1-4}
\rowcolor{gray!10}
Full Token & 50.7 & 56.3 & -- \\
Fixed segmentation  & 49.3 & 54.3 & \textbf{56} \\
Depth-based pruning & 49.2 & 54.3 & 54 \\
Depth-based merging  & \textbf{50.6} & \textbf{55.5} & 54 \\
\bottomrule
\end{tabular}
\vspace{-2mm}
\end{minipage}
\hfill
\begin{minipage}[t]{0.49\linewidth}
\centering
\setlength{\tabcolsep}{3.2pt}
\renewcommand{\arraystretch}{0.95}
\caption{\textbf{Efficiency--accuracy comparison.}
Comparison across compression (\%), memory (GB), prefill time (s), latency (s), and accuracy.}
\label{tab:efficiency_wrap}
\begin{tabular}{lccccc}
\toprule
Method & Comp. & Mem.$\downarrow$ & Pre.$\downarrow$ & Lat.$\downarrow$ & Acc.$\uparrow$ \\
\midrule
\multicolumn{6}{c}{\textit{Qwen2.5-Omni 7B}} \\
\cmidrule(lr){1-6}
\rowcolor{gray!10}
Full Token & - & 27.1 & 5.0 & 6.7 & \textbf{60.8} \\
FastV & 48 & \textbf{23.8} & 3.1 & 4.7 & 60.4 \\
OmniZip & 53 & 24.0 & \textbf{2.9} & 4.7 & 59.4 \\
Ours & \textbf{54} & 24.6 & 3.1 & \textbf{4.5} & \textbf{60.8} \\
\midrule
\multicolumn{6}{c}{\textit{Video-SALMONN2+ 7B}} \\
\cmidrule(lr){1-6}
\rowcolor{gray!10}
Full Token & - & 22.5 & 4.4 & 4.5 & \textbf{62.0} \\
FastV & 49 & \textbf{20.3} & 4.1 & 4.1 & 59.4 \\
OmniZip & 48 & 20.4 & \textbf{3.8} & \textbf{3.9} & 60.2 \\
Ours & \textbf{53} & 20.8 & 3.9 & \textbf{3.9} & \textbf{62.0} \\
\bottomrule
\end{tabular}
\end{minipage}
\vspace{-3mm}
\end{table}

\subsection{Efficiency analysis}
Table~\ref{tab:efficiency_wrap} reports inference statistics on 100 samples from each of the five non-captioning benchmarks for Qwen2.5-Omni 7B and Video-SALMONN2+ 7B, covering peak GPU memory usage (Mem.), prefill time (Pre.), and end-to-end latency (Lat.). 
Our method reduces memory usage relative to full-token inference and, despite the slight overhead from the A2V predictor, achieves or matches the lowest latency among inference-time pruning baselines while preserving full-token-level accuracy.

\subsection{A chunkwise online-friendly variant}
\begin{wraptable}{r}{0.43\textwidth}
\vspace{-1em}
\centering
\footnotesize
\setlength{\tabcolsep}{2.6pt}
\renewcommand{\arraystretch}{0.95}
\vspace{-1mm}
\caption{\textbf{Chunkwise online-friendly variant.} We compare the online-friendly variant with the main offline method.}
\vspace{-1mm}
\begin{tabular}{lccc}
\toprule
Method & World. & Daily. & Comp. (\%) \\
\midrule
\multicolumn{4}{c}{\textit{Qwen2.5-Omni 7B}} \\
\cmidrule(lr){1-4}
\rowcolor{gray!10}
Full Token & 47.4 & 57.1 & -- \\
Ours (online) & {47.1} & 56.1  & 50 \\
Ours (offline) & \textbf{47.7} & \textbf{57.2}  & \textbf{52} \\
\midrule
\multicolumn{4}{c}{\textit{Video-SALMONN2+ 7B}} \\
\cmidrule(lr){1-4}
\rowcolor{gray!10}
Full Token & 50.7 & 56.3 & -- \\
Ours (online) & \textbf{51.2} & {55.4}  & 50 \\
Ours (offline) & {50.6} & \textbf{55.5}  & \textbf{54} \\
\bottomrule
\end{tabular}
\label{tab:online_variant}
\vspace{-6mm}
\end{wraptable}
Although our main method uses offline depth-based merging, the same chunkwise pruning principle suggests a simple online-friendly variant. In this variant, temporal compression relies only on local chunk-to-chunk similarity. We evaluate this variant on Qwen2.5-Omni 7B and Video-SALMONN2+ 7B on the World. and Daily. benchmarks.

As shown in Table~\ref{tab:online_variant}, the online-friendly variant remains competitive with the main offline method at around 50\% compression ratio. These results suggest that the proposed pruning principle may extend to an online-friendly setting while maintaining competitive performance. Further details are provided in App.~\ref{supp:online_variant}.

\section{Related Work}

\paragraph{Omni-LLMs.}
Recent progress in multimodal large language models (MLLMs)~\cite{alayrac2022flamingo, chen2023shikra,huang2023language,li2022blip,maaz2023video, touvron2023llama,yu2024rlhf,zhang2023llama, zhu2023minigpt} has extended beyond vision-language understanding to jointly model vision, audio, and text~\cite{chen2023vast, cheng2024videollama, chowdhury2024meerkat, han2024onellm, lyu2023macaw, panagopoulou2023x, ye2024cat, zhan2024anygpt, zhang2023video, zhao2023chatbridge}. 
Proprietary systems such as GPT-4o~\cite{openai2024gpt4ocard} and the Gemini series~\cite{comanici2025gemini25pushingfrontier,geminiteam2024gemini15unlockingmultimodal} demonstrate strong capabilities on AV reasoning, while the open-source community has produced a growing line of models~\cite{sun2024video,tang2025video,qwen25omni}. 
A prevalent architectural paradigm couples modality-specific encoders~\cite{beats2023, dosovitskiy2020image,gong2021astaudiospectrogramtransformer, radford2021learning,whisper} with an LLM backbone, projecting modality-specific tokens into a shared embedding space and interleaving them into chunk-structured sequences. 
While effective, this design introduces a severe computational bottleneck. Even a short clip produces thousands of tokens, and the quadratic cost of self-attention makes efficient inference essential for practical deployment.
Token compression, which reduces multimodal tokens either before entering the LLM or within its early layers, has emerged as a promising alternative to address this problem.

\paragraph{Token Compression for MLLMs.}
To reduce the computational cost of visual processing in MLLMs, many video token 
compression methods~\cite{chen2024image,chen2026streamingtom,hyun2025multi,
liu2025mixkv,qi2026adapttoken,shang2025llavaprumerge,shao2025holitom,
tan2025tokencarve,tao2025dycoke,xing2025pyramiddropacceleratinglargevisionlanguage,
yang2024pvcprogressivevisualtoken,yang2026visionziplongerbetternecessary,
yao2025timechatonline} prune spatio-temporally redundant tokens using 
similarity or saliency. While effective, these methods focus on video-only settings and cannot exploit the cross-modal structure of joint AV streams. 
Recent works~\cite{gong2025echoingpixels, ding2026omnisift} have begun to 
study token compression in Omni-LLMs, and can be categorized by whether 
they require downstream supervision. 
EchoingPixels~\cite{gong2025echoingpixels} jointly compresses audio and 
video tokens via a bidirectional cross-modal encoder with redesigned 
positional encoding to preserve temporal relationships after pruning.
OmniSIFT~\cite{ding2026omnisift} prunes video tokens by spatio-temporal saliency, then uses the retained 
visual anchors to guide audio selection through a trainable cross-attention 
module. In contrast, OmniZip~\cite{zhao2026omnizip} proposes a 
training-free framework that identifies salient audio tokens and leverages 
their retention patterns to guide video token compression. 
These methods generally frame pruning as selecting the most informative subset of tokens, guided by downstream supervision or strongly aligned AV events. In contrast, we approach AV token compression from a context-preservation perspective, removing audio-predictable coarse visual semantics while preserving non-audio-aligned evidence and localized visual details.
Furthermore, ContextGuard operates without downstream supervision, 
making OmniZip the most directly comparable prior method. 
This distinction is important in realistic AV interaction, where information outside the current query or strongly aligned regions may still be needed for broad context understanding.
\section{Conclusion}
We presented ContextGuard, an audio-guided token pruning framework that reframes Omni-LLM token reduction from selecting important tokens to removing cross-modal redundancy while preserving broad AV context.
ContextGuard combines audio-guided semantic pruning, spatial detail preservation, and depth-based merging to reduce token usage while maintaining downstream performance across multiple backbones and benchmarks.
On Qwen2.5-Omni 7B, it achieves full-token-level performance on five of six benchmarks while pruning 55\% of input tokens.

\clearpage
\bibliography{shorstrings,main}{}

@inproceedings{alayrac2022flamingo,
    author ={Alayrac, Jean-Baptiste and Donahue, Jeff and Luc, Pauline and Miech, Antoine and Barr, Iain and Hasson, Yana and Lenc, Karel and Mensch, Arthur and Millican, Katherine and Reynolds, Malcolm and others} ,
    title ={{Flamingo: a Visual Language Model for Few-Shot Learning}} ,
    booktitle =NIPS,
    year = {2022} 
}

@article{bai2025qwen25vltechnicalreport,
      title={{Qwen2.5-VL Technical Report}}, 
      author={Shuai Bai and Keqin Chen and Xuejing Liu and Jialin Wang and Wenbin Ge and Sibo Song and Kai Dang and Peng Wang and Shijie Wang and Jun Tang and others},
      year={2025},
      journal={arXiv},
}

@inproceedings{chen2020vggsound,
  title={{VGGSound: A Large-scale Audio-Visual Dataset}},
  author={Chen, Honglie and Xie, Weidi and Vedaldi, Andrea and Zisserman, Andrew},
  booktitle=ICASSP,
  year={2020},
}

@article{chen2023shikra,
  title={{Shikra: Unleashing Multimodal LLM’s Referential Dialogue Magic}},
  author={Chen, Keqin and Zhang, Zhao and Zeng, Weili and Zhang, Richong and Zhu, Feng and Zhao, Rui},
  journal={arXiv:2306.15195},
  year={2023}
}

@inproceedings{chen2024image,
  title={{An Image is Worth 1/2 Tokens After Layer 2: Plug-and-Play Inference Acceleration for Large Vision-Language Models}},
  author={Chen, Liang and Zhao, Haozhe and Liu, Tianyu and Bai, Shuai and Lin, Junyang and Zhou, Chang and Chang, Baobao},
  booktitle=ECCV,
  year={2024},
}

@InProceedings{beats2023,
  title = 	 {{BEATs: Audio Pre-Training with Acoustic Tokenizers}},
  author =       {Chen, Sanyuan and Wu, Yu and Wang, Chengyi and Liu, Shujie and Tompkins, Daniel and Chen, Zhuo and Che, Wanxiang and Yu, Xiangzhan and Wei, Furu},
  booktitle =ICML,
  year = 	 {2023}
}

@inproceedings{chen2023vast,
    author = {Chen, Sihan and Li, Handong and Wang, Qunbo and Zhao, Zijia and Sun, Mingzhen and Zhu, Xinxin and Liu, Jing},
    title ={{VAST: A Vision-Audio-Subtitle-Text Omni-Modality Foundation Model and Dataset}},
    booktitle =nips ,
    year = {2023}
}

@inproceedings{chen2026streamingtom,
  title={{StreamingTOM: Streaming Token Compression for Efficient Video Understanding}},
  author={Chen, Xueyi and Tao, Keda and Shao, Kele and Wang, Huan},
  booktitle=CVPR,
  year={2026}
}

@inproceedings{chen2024internvl,
    title={{InternVL: Scaling Up Vision Foundation Models and Aligning for Generic Visual-Linguistic Tasks}},
    author={Chen, Zhe and Wu, Jiannan and Wang, Wenhai and Su, Weijie and Chen, Guo and Xing, Sen and Zhong, Muyan and Zhang, Qinglong and Zhu, Xizhou and Lu, Lewei and others},
    booktitle=CVPR,
    year={2024}
  }

@article{cheng2024videollama,
  title={{VideoLLaMA 2: Advancing Spatial-Temporal Modeling and Audio Understanding in Video-LLMs}},
  author={Cheng, Zesen and Leng, Sicong and Zhang, Hang and Xin, Yifei and Li, Xin and Chen, Guanzheng and Zhu, Yongxin and Zhang, Wenqi and Luo, Ziyang and Zhao, Deli and others},
  journal={arXiv:2406.07476},
  year={2024},
}

@inproceedings{chowdhury2024meerkat,
  title={{Meerkat: Audio-Visual Large Language Model for Grounding in Space and Time}},
  author={Chowdhury, Sanjoy and Nag, Sayan and Dasgupta, Subhrajyoti and Chen, Jun and Elhoseiny, Mohamed and Gao, Ruohan and Manocha, Dinesh},
  booktitle=ECCV,
  year={2024},
}

@article{comanici2025gemini25pushingfrontier,
      title={{Gemini 2.5: Pushing the Frontier with Advanced Reasoning, Multimodality, Long Context, and Next Generation Agentic Capabilities}}, 
      author={Gheorghe Comanici and Eric Bieber and Mike Schaekermann and Ice Pasupat and Noveen Sachdeva and Inderjit Dhillon and Marcel Blistein and Ori Ram and Dan Zhang and Evan Rosen and others},
      year={2025},
      journal={arXiv:2507.06261},
}

@article{ding2026omnisift,
  title={{OmniSIFT: Modality-Asymmetric Token Compression for Efficient Omni-modal Large Language Models}},
  author={Ding, Yue and Ji, Yiyan and Li, Jungang and Liu, Xuyang and Chen, Xinlong and Wu, Junfei and Li, Bozhou and Zeng, Bohan and Shi, Yang and Guan, Yushuo and others},
  journal={arXiv:2602.04804},
  year={2026}
}

@inproceedings{dosovitskiy2020image,
    author = {Alexey Dosovitskiy and Lucas Beyer and Alexander Kolesnikov and Dirk Weissenborn and Xiaohua Zhai and Thomas Unterthiner and Mostafa Dehghani and Matthias Minderer and Georg Heigold and Sylvain Gelly and Jakob Uszkoreit and Neil Houlsby},
    title = {{An Image is Worth 16x16 Words: Transformers for Image Recognition at Scale}},
    booktitle = ICLR,
    year = 2021
}

@inproceedings{fu2025video,
  title={{Video-MME: The First-Ever Comprehensive Evaluation Benchmark of Multi-modal LLMs in Video Analysis}},
  author={Fu, Chaoyou and Dai, Yuhan and Luo, Yongdong and Li, Lei and Ren, Shuhuai and Zhang, Renrui and Wang, Zihan and Zhou, Chenyu and Shen, Yunhang and Zhang, Mengdan and others},
  booktitle=CVPR,
  year={2025}
}

@inproceedings{gemmeke2017audio,
  title={{Audio Set: An ontology and human-labeled dataset for audio events}},
  author={Gemmeke, Jort F. and Ellis, Daniel P. W. and Freedman, Dylan and Jansen, Aren and Lawrence, Wade and Moore, R. Channing and Plakal, Manoj and Ritter, Marvin},
  booktitle=ICASSP,
  year={2017}
}

@article{gong2025echoingpixels,
  title={{EchoingPixels: Cross-Modal Adaptive Token Reduction for Efficient Audio-Visual LLMs}},
  author={Gong, Chao and Wang, Depeng and Wei, Zhipeng and Guo, Ya and Zhu, Huijia and Chen, Jingjing},
  journal={arXiv:2512.10324},
  year={2025}
}

@inproceedings{gong2021astaudiospectrogramtransformer,
      title={{AST: Audio Spectrogram Transformer}}, 
      author={Yuan Gong and Yu-An Chung and James Glass},
      year={2021},
      booktitle=is
}

@inproceedings{han2024onellm,
  title={{OneLLM: One Framework to Align All Modalities with Language}},
  author={Han, Jiaming and Gong, Kaixiong and Zhang, Yiyuan and Wang, Jiaqi and Zhang, Kaipeng and Lin, Dahua and Qiao, Yu and Gao, Peng and Yue, Xiangyu},
  booktitle=CVPR,
  year={2024}
}

@inproceedings{hong2025worldsense,
    author = {Hong, Jack and Yan, Shilin and Cai, Jiayin and Jiang, Xiaolong and Hu, Yao and Xie, Weidi},
    title ={{WorldSense: Evaluating Real-world Omnimodal Understanding for Multimodal LLMs}} ,
    booktitle =ICLR ,
    year = {2026}
}

@inproceedings{huang2023language,
    author ={Huang, Shaohan and Dong, Li and Wang, Wenhui and Hao, Yaru and Singhal, Saksham and Ma, Shuming and Lv, Tengchao and Cui, Lei and Mohammed, Owais Khan and Patra, Barun and others} ,
    title = {{Language is Not All You Need: Aligning Perception with Language Models}},
    booktitle =NIPS,
    year = {2023}
}

@article{openai2024gpt4ocard,
      title={{GPT-4o System Card}}, 
      author={Aaron Hurst and Adam Lerer and Adam P. Goucher and Adam Perelman and Aditya Ramesh and Aidan Clark and AJ Ostrow and Akila Welihinda and Alan Hayes and Alec Radford and others},
      year={2024},
      journal={arXiv:2410.21276},
}

@inproceedings{hyun2025multi,
  title={{Multi-Granular Spatio-Temporal Token Merging for Training-Free Acceleration of Video LLMs}},
  author={Hyun, Jeongseok and Hwang, Sukjun and Han, Su Ho and Kim, Taeoh and Lee, Inwoong and Wee, Dongyoon and Lee, Joon-Young and Kim, Seon Joo and Shim, Minho},
  booktitle=ICCV,
  year={2025}
}

@inproceedings{jiang2025stormtokenefficientlongvideo,
      title={{STORM: Token-Efficient Long Video Understanding for Multimodal LLMs}}, 
      author={Jindong Jiang and Xiuyu Li and Zhijian Liu and Muyang Li and Guo Chen and Zhiqi Li and De-An Huang and Guilin Liu and Zhiding Yu and Kurt Keutzer and Sungjin Ahn and Jan Kautz and Hongxu Yin and Yao Lu and Song Han and Wonmin Byeon},
      year={2025},
      booktitle=iccvwork,
}

@article{lillava,
  title={{LLaVA-OneVision: Easy Visual Task Transfer}},
  author={Li, Bo and Zhang, Yuanhan and Guo, Dong and Zhang, Renrui and Li, Feng and Zhang, Hao and Zhang, Kaichen and Zhang, Peiyuan and Li, Yanwei and Liu, Ziwei and Li, Chunyuan},
  journal={Transactions on Machine Learning Research},
  year={2024}
}

@article{li2025omnivideobench,
  title={{OmniVideoBench: Towards Audio-Visual Understanding Evaluation for Omni MLLMs}},
  author={Li, Caorui and Chen, Yu and Ji, Yiyan and Xu, Jin and Cui, Zhenyu and Li, Shihao and Zhang, Yuanxing and Wang, Wentao and Song, Zhenghao and Zhang, Dingling and others},
  journal={arXiv:2510.10689},
  year={2025}
}

@inproceedings{li2022blip,
  title={{BLIP-2: Bootstrapping Language-Image Pre-training with Frozen Image Encoders and Large Language Models}},
  author={Li, Junnan and Li, Dongxu and Savarese, Silvio and Hoi, Steven},
  booktitle=ICML,
  year={2023}
}

@inproceedings{li2024videochat,
  title={{VideoChat-Flash: Hierarchical Compression for Long-Context Video Modeling}},
  author={Li, Xinhao and Wang, Yi and Yu, Jiashuo and Zeng, Xiangyu and Zhu, Yuhan and Huang, Haian and Gao, Jianfei and Li, Kunchang and He, Yinan and Wang, Chenting and Qiao, Yu and Wang, Yali and Wang, Limin},
  booktitle=ICLR,
  year={2026}
}

@inproceedings{lin-etal-2024-video,
  author    = {Lin, Bin and Ye, Yang and Zhu, Bin and Cui, Jiaxi and Ning, Munan and Jin, Peng and Yuan, Li},
  title     = {{Video-LLaVA: Learning United Visual Representation by Alignment Before Projection}},
  booktitle = emnlp,
  year      = {2024}
}

@inproceedings{liu2023llava,
    author ={Liu, Haotian and Li, Chunyuan and Wu, Qingyang and Lee, Yong Jae} ,
    title ={{Visual Instruction Tuning}},
    booktitle =NIPS,
    year = {2023}
}

@inproceedings{liu2025mixkv,
    author = {Liu, Xuyang and Gui, Xiyan and Zhang, Yuchao and Zhang, Linfeng},
    title={{Mixing Importance with Diversity: Joint Optimization for KV Cache Compression in Large Vision-Language Models}},
    booktitle = ICLR,
    year = {2026}
}

@article{lyu2023macaw,
  title={{Macaw-LLM: Multi-Modal Language Modeling with Image, Audio, Video, and Text Integration}},
  author={Lyu, Chenyang and Wu, Minghao and Wang, Longyue and Huang, Xinting and Liu, Bingshuai and Du, Zefeng and Shi, Shuming and Tu, Zhaopeng},
  journal={arXiv:2306.09093},
  year={2023}
}

@inproceedings{maaz2023video,
    author ={Maaz, Muhammad and Rasheed, Hanoona and Khan, Salman and Khan, Fahad Shahbaz} ,
    title ={{Video-ChatGPT: Towards Detailed Video Understanding via Large Vision and Language Models}} ,
    booktitle = ACL,
    year = {2024}
}

@article{panagopoulou2023x,
  title={{X-InstructBLIP: A Framework for Aligning X-Modal Instruction-Aware Representations to LLMs and Emergent Cross-modal Reasoning}},
  author={Panagopoulou, Artemis and Xue, Le and Yu, Ning and Li, Junnan and Li, Dongxu and Joty, Shafiq and Xu, Ran and Savarese, Silvio and Xiong, Caiming and Niebles, Juan Carlos},
  journal={arXiv:2311.18799},
  year={2023}
}

@article{qi2026adapttoken,
      title={{AdaptToken: Entropy-based Adaptive Token Selection for MLLM Long Video Understanding}}, 
      author={Haozhe Qi and Kevin Qu and Mahdi Rad and Rui Wang and Alexander Mathis and Marc Pollefeys},
      year={2026},
      journal={arXiv:2603.28696},
}

@article{qi2025lmm,
  title={{Quicksviewer: An LMM for Efficient Video Understanding via Reinforced Compression of Video Cubes}},
  author={Qi, Ji and Yao, Yuan and Bai, Yushi and Xu, Bin and Li, Juanzi and Liu, Zhiyuan and Chua, Tat-Seng},
  journal={arXiv:2504.15270},
  year={2025}
}

@inproceedings{radford2021learning,
  title={{Learning Transferable Visual Models From Natural Language Supervision}},
  author={Radford, Alec and Kim, Jong Wook and Hallacy, Chris and Ramesh, Aditya and Goh, Gabriel and Agarwal, Sandhini and Sastry, Girish and Askell, Amanda and Mishkin, Pamela and Clark, Jack and others},
  booktitle=ICML,
  year={2021},
}

@inproceedings{whisper,
author = {Radford, Alec and Kim, Jong Wook and Xu, Tao and Brockman, Greg and McLeavey, Christine and Sutskever, Ilya},
title = {{Robust Speech Recognition via Large-Scale Weak Supervision}},
year = {2023},
booktitle = ICML
}

@inproceedings{shang2025llavaprumerge,
  title={{LLaVA-PruMerge: Adaptive Token Reduction for Efficient Large Multimodal Models}},
  author={Shang, Yuzhang and Cai, Mu and Xu, Bingxin and Lee, Yong Jae and Yan, Yan},
  booktitle=ICCV,
  year={2025}
}

@ARTICLE{6773024,
  author={Shannon, C. E.},
  journal={The Bell System Technical Journal}, 
  title={A mathematical theory of communication}, 
  year={1948},
  volume={27},
  number={3},
  pages={379-423},
  doi={10.1002/j.1538-7305.1948.tb01338.x}
  }

@inproceedings{shao2025holitom,
  title={{HoliTom: Holistic Token Merging for Fast Video Large Language Models}},
  author={Shao, Kele and Tao, Keda and Qin, Can and You, Haoxuan and Sui, Yang and Wang, Huan},
  booktitle=NIPS,
  year={2025}
}

@inproceedings{shu2025video,
  title={{Video-XL: Extra-Long Vision Language Model for Hour-Scale Video Understanding}},
  author={Shu, Yan and Liu, Zheng and Zhang, Peitian and Qin, Minghao and Zhou, Junjie and Liang, Zhengyang and Huang, Tiejun and Zhao, Bo},
  booktitle=CVPR,
  year={2025}
}

@inproceedings{sun2024video,
  title={{video-SALMONN: Speech-Enhanced Audio-Visual Large Language Models}},
  author={Sun, Guangzhi and Yu, Wenyi and Tang, Changli and Chen, Xianzhao and Tan, Tian and Li, Wei and Lu, Lu and Ma, Zejun and Wang, Yuxuan and Zhang, Chao},
  booktitle=icml,
  year={2024}
}

@article{tan2025tokencarve,
  title={{TokenCarve: Information-Preserving Visual Token Compression in Multimodal Large Language Models}},
  author={Tan, Xudong and Ye, Peng and Tu, Chongjun and Cao, Jianjian and Yang, Yaoxin and Zhang, Lin and Zhou, Dongzhan and Chen, Tao},
  year={2025},
  journal={arXiv:2503.10501},
}

@article{tang2025video,
  title={{video-SALMONN 2: Caption-Enhanced Audio-Visual Large Language Models}},
  author={Tang, Changli and Li, Yixuan and Yang, Yudong and Zhuang, Jimin and Sun, Guangzhi and Li, Wei and Ma, Zejun and Zhang, Chao},
  journal={arXiv:2506.15220},
  year={2025}
}

@inproceedings{tao2025dycoke,
  title={{DyCoke: Dynamic Compression of Tokens for Fast Video Large Language Models}},
  author={Tao, Keda and Qin, Can and You, Haoxuan and Sui, Yang and Wang, Huan},
  booktitle=CVPR,
  year={2025}
}

@inproceedings{zhao2026omnizip,
      title={{OmniZip: Audio-Guided Dynamic Token Compression for Fast Omnimodal Large Language Models}}, 
      author={Keda Tao and Kele Shao and Bohan Yu and Weiqiang Wang and Jian Liu and Huan Wang},
      year={2026},
      booktitle=CVPR, 
}

@article{geminiteam2024gemini15unlockingmultimodal,
      title={Gemini 1.5: Unlocking multimodal understanding across millions of tokens of context}, 
      author={Gemini Team and Petko Georgiev and Ving Ian Lei and Ryan Burnell and Libin Bai and Anmol Gulati and Garrett Tanzer and Damien Vincent and Zhufeng Pan and Shibo Wang and Soroosh Mariooryad and Yifan Ding and Xinyang Geng and Fred Alcober and Roy Frostig and Mark Omernick and Lexi Walker and others},
      year={2024},
      journal={arXiv:2403.05530},
}

@article{touvron2023llama,
  title={{LLaMA: Open and Efficient Foundation Language Models}},
  author={Touvron, Hugo and Lavril, Thibaut and Izacard, Gautier and Martinet, Xavier and Lachaux, Marie-Anne and Lacroix, Timoth{\'e}e and Rozi{\`e}re, Baptiste and Goyal, Naman and Hambro, Eric and Azhar, Faisal and others},
  journal={arXiv:2302.13971},
  year={2023}
}

@inproceedings{xing2025pyramiddropacceleratinglargevisionlanguage,
    author = {Long Xing and Qidong Huang and Xiaoyi Dong and Jiajie Lu and Pan Zhang and Yuhang Zang and Yuhang Cao and Conghui He and Jiaqi Wang and Feng Wu and Dahua Lin},
    title = {{PyramidDrop: Accelerating Your Large Vision-Language Models via Pyramid Visual Redundancy Reduction}},
    booktitle = CVPR,
    year = 2025
}

@article{qwen25omni,
  author  = {Xu, Jin and Guo, Zhifang and He, Jinzheng and Hu, Hangrui and He, Ting and Bai, Shuai and Chen, Keqin and Wang, Jialin and Fan, Yang and Dang, Kai and others},
  title   = {{Qwen2.5-Omni Technical Report}},
  journal={arXiv:2503.20215},
  year    = {2025}
}

@inproceedings{yang2024pvcprogressivevisualtoken,
      title={{PVC: Progressive Visual Token Compression for Unified Image and Video Processing in Large Vision-Language Models}}, 
      author={Chenyu Yang and Xuan Dong and Xizhou Zhu and Weijie Su and Jiahao Wang and Hao Tian and Zhe Chen and Wenhai Wang and Lewei Lu and Jifeng Dai},
      booktitle=CVPR,
    year={2025}
}

@inproceedings{yang2022avqa,
  title={{AVQA: A Dataset for Audio-Visual Question Answering on Videos}},
  author={Yang, Pinci and Wang, Xin and Duan, Xuguang and Chen, Hong and Hou, Runze and Jin, Cong and Zhu, Wenwu},
  booktitle=acmmm,
  year={2022}
}

@inproceedings{yang2026visionziplongerbetternecessary,
      title={{VisionZip: Longer is Better but Not Necessary in Vision Language Models}}, 
      author={Senqiao Yang and Yukang Chen and Zhuotao Tian and Chengyao Wang and Jingyao Li and Bei Yu and Jiaya Jia},
      year={2025},
      booktitle=CVPR,
}

@inproceedings{ye2024cat,
  title={{CAT: Enhancing Multimodal Large Language Model to Answer Questions in Dynamic Audio-Visual Scenarios}},
  author={Ye, Qilang and Yu, Zitong and Shao, Rui and Xie, Xinyu and Torr, Philip and Cao, Xiaochun},
  booktitle=ECCV,
  year={2024},
}

@inproceedings{ye2024fitprunefasttrainingfree,
      title={{Fit and Prune: Fast and Training-free Visual Token Pruning for Multi-modal Large Language Models}}, 
      author={Weihao Ye and Qiong Wu and Wenhao Lin and Yiyi Zhou},
      year={2025},
      booktitle=aaai
}

@inproceedings{yu2024rlhf,
    author = {Yu, Tianyu and Yao, Yuan and Zhang, Haoye and He, Taiwen and Han, Yifeng and Cui, Ganqu and Hu, Jinyi and Liu, Zhiyuan and Zheng, Hai-Tao and Sun, Maosong and others},
    title = {{RLHF-V: Towards Trustworthy MLLMs via Behavior Alignment from Fine-grained Correctional Human Feedback}},
    booktitle = CVPR,
    year = {2024}
}

@article{zhan2024anygpt,
  title={{AnyGPT: Unified Multimodal LLM with Discrete Sequence Modeling}},
  author={Zhan, Jun and Dai, Junqi and Ye, Jiasheng and Zhou, Yunhua and Zhang, Dong and Liu, Zhigeng and Zhang, Xin and Yuan, Ruibin and Zhang, Ge and Li, Linyang and others},
  journal={arXiv:2402.12226},
  year={2024}
}

@inproceedings{zhang2023video,
  author    = {Zhang, Hang and Li, Xin and Bing, Lidong},
  title     = {{Video-LLaMA: An Instruction-tuned Audio-Visual Language Model for Video Understanding}},
  booktitle = emnlp,
  year      = {2023}
}

@inproceedings{zhang2024pmod,
  title={{p-MoD: Building Mixture-of-Depths MLLMs via Progressive Ratio Decay}},
  author={Zhang, Jun and Meng, Desen and Qi, Ji and Huang, Zhenpeng and Wu, Tao and Wang, Limin},
  booktitle=ICCV,
  year={2025}
}

@inproceedings{zhang2023llama,
    author ={Zhang, Renrui and Han, Jiaming and Liu, Chris and Gao, Peng and Zhou, Aojun and Hu, Xiangfei and Yan, Shilin and Lu, Pan and Li, Hongsheng and Qiao, Yu} ,
    title = {{LLaMA-Adapter: Efficient Fine-tuning of Language Models with Zero-init Attention}},
    booktitle = ICLR,
    year = {2024}
}

@article{zhao2023chatbridge,
  title={{ChatBridge: Bridging Modalities with Large Language Model as a Language Catalyst}},
  author={Zhao, Zijia and Guo, Longteng and Yue, Tongtian and Chen, Sihan and Shao, Shuai and Zhu, Xinxin and Yuan, Zehuan and Liu, Jing},
  journal={arXiv:2305.16103},
  year={2023}
}

@article{zhou2025daily,
  title={{Daily-Omni: Towards Audio-Visual Reasoning with Temporal Alignment across Modalities}},
  author={Zhou, Ziwei and Wang, Rui and Wu, Zuxuan and Jiang, Yu-Gang},
  journal={arXiv:2505.17862},
  year={2025}
}

@inproceedings{zhu2023minigpt,
    author ={Zhu, Deyao and Chen, Jun and Shen, Xiaoqian and Li, Xiang and Elhoseiny, Mohamed} ,
    title ={{MiniGPT-4: Enhancing Vision-Language Understanding with Advanced Large Language Models}},
    booktitle = ICLR,
    year = {2024}
}

@inproceedings{yao2025timechatonline,
  title={{TimeChat-Online: 80\% Visual Tokens are Naturally Redundant in Streaming Videos}},
  author={Yao, Linli and Li, Yicheng and Wei, Yuancheng and Li, Lei and Ren, Shuhuai and Liu, Yuanxin and Ouyang, Kun and Wang, Lean and Li, Shicheng and Li, Sida and others},
  booktitle=ACMMM,
  year={2025}
}

@string{aaai =  "Proc. AAAI"}

@string{acmmm = "Proc. ACM MM"}

@string{acl = "Proc. ACL"}

@string{cvpr   = "Proc. CVPR"}

@string{eccv  = "Proc. ECCV"}

@string{emnlp  = "Proc. EMNLP"}

@string{icassp="Proc. ICASSP"}

@string{iccv =  "Proc. ICCV"}

@string{iccvwork =  "Proc. ICCV Workshop"}

@string{iclr = "Proc. ICLR"}

@string{icml =  "Proc. ICML"}

@string{is =  "Proc. Interspeech"}

@string{nips = "Proc. NeurIPS"}
\bibliographystyle{plain}


\clearpage

\appendix

\section*{Appendix}

\section{Audio-to-Video Semantic Predictor}
\label{supp:A}

\subsection{Architecture and Training Details}

\label{supp:a2v_predictor}

\paragraph{Predictor architecture and training objective.}
We implement the audio-to-video semantic predictor (A2V predictor) as a lightweight module with two cross-attention layers and 128 learnable queries.
The predictor maps aligned audio token features into the visual embedding space through cross-attention and a lightweight MLP head.
Supervision is applied only at the global semantic level: both the predicted token sequence and the target visual token sequence are mean-pooled into a single semantic vector, and the losses are computed between these pooled representations.
This encourages the predictor to capture coarse audio-shared visual semantics rather than patch-level details.

We use multiple learnable queries to provide the predictor with sufficient capacity to summarize diverse audio-implied visual aspects within a chunk. A single query could in principle produce one global audio-to-visual summary, but multiple queries form a richer bottleneck before aggregation. We mean-pool the query outputs because ContextGuard only requires a stable chunk-level semantic prototype for estimating audio explainability, rather than query-specific patch-level predictions.

Let $\hat{\bar{\mathbf{h}}}^{v}_{b,t}$ denote the mean-pooled visual-semantic prediction produced from the audio tokens of chunk $t$ in video $b$, and let $\bar{\mathbf{h}}^{v}_{b,t}$ denote the mean-pooled target visual representation of the corresponding video chunk.
Given a batch of $B$ videos with $T$ chunks each, we train the predictor with a cosine alignment loss and a contrastive loss:
\begin{equation}
\begin{split}
\mathcal{L}_{\mathrm{cos}}
&=
\frac{1}{BT}\sum_{b=1}^{B}\sum_{t=1}^{T}
\left(
1-\mathrm{sim}\!\left(
\hat{\bar{\mathbf{h}}}^{v}_{b,t},
\bar{\mathbf{h}}^{v}_{b,t}
\right)
\right), \\
\mathcal{L}_{\mathrm{ctr}}
&=
-\frac{1}{BT}\sum_{b=1}^{B}\sum_{t=1}^{T}
\log
\frac{
\exp\!\left(
\mathrm{sim}(\hat{\bar{\mathbf{h}}}^{v}_{b,t}, \bar{\mathbf{h}}^{v}_{b,t})/\tau
\right)
}{
\exp\!\left(
\mathrm{sim}(\hat{\bar{\mathbf{h}}}^{v}_{b,t}, \bar{\mathbf{h}}^{v}_{b,t})/\tau
\right)
+
\sum\limits_{\substack{b'=1, \\ b'\neq b}}^{B}
\sum\limits_{t'=1}^{T}
\exp\!\left(
\mathrm{sim}(\hat{\bar{\mathbf{h}}}^{v}_{b,t}, \bar{\mathbf{h}}^{v}_{b',t'})/\tau
\right)
}.
\end{split}
\label{eq:semantic_losses}
\end{equation}
For each prediction $\hat{\bar{\mathbf{h}}}^{v}_{b,t}$, the matched target $\bar{\mathbf{h}}^{v}_{b,t}$ is used as the positive.
The contrastive loss uses visual targets from other videos in the same batch as negatives, while excluding other chunks from the same video to avoid false negatives from temporally adjacent chunks with similar semantics.

The total training objective is
\begin{equation}
\mathcal{L}_{\mathrm{sem}}
=
\lambda_{\mathrm{cos}}\mathcal{L}_{\mathrm{cos}}
+
\mathcal{L}_{\mathrm{ctr}},
\label{eq:semantic_training_loss}
\end{equation}
with $\lambda_{\mathrm{cos}}=5.0$ and contrastive temperature $\tau=0.07$.

\paragraph{Training data and setup for the predictor.}
The A2V predictor is trained on a mixture of AudioSet~\cite{gemmeke2017audio} and the VGGSound~\cite{chen2020vggsound} training split. 
We train the predictor using a batch size of 8 with gradient accumulation over 2 steps on 4 GPUs. 
In practice, the model converges early, typically within 10k–15k training steps. 

\paragraph{Why use an A2V predictor for semantic scoring?}

\begin{wraptable}{r}{0.3\textwidth}
\vspace{-1.0em}
\centering
\footnotesize
\setlength{\tabcolsep}{3.0pt}
\renewcommand{\arraystretch}{0.92}
\caption{\textbf{Reference embeddings for semantic scoring.}
A2V gives the strongest performance on Qwen2.5-Omni.}
\vspace{-2mm}
\begin{tabular}{lcc}
\toprule
Method & World. & Daily.  \\
\midrule
\multicolumn{3}{c}{\textit{Qwen2.5-Omni 7B}} \\
\cmidrule(lr){1-3}
orig & 47.6 & 56.7 \\
video mean & 47.1 & 56.0  \\
ours & \textbf{47.7} & \textbf{57.2} \\
\bottomrule
\end{tabular}
\label{supp:tab:sem_scoring}
\vspace{-1em}
\end{wraptable}
We compare semantic redundancy scores computed with the original audio embedding, the A2V-predicted visual-semantic embedding, and the mean visual embedding.
The original audio embedding depends on the backbone's native audio-video alignment: it can work when alignment is strong, but becomes unreliable when alignment is weak, as shown in Table~\ref{tab:retrieval_predictor}.
The mean visual embedding forms a collapsed visual prototype and therefore scores visual deviation from the chunk average rather than cross-modal redundancy with audio.
These results support the A2V predictor as a more reliable reference for semantic redundancy scoring.

\subsection{Audio-to-Video Retrieval Analysis}
\label{supp:a2v_retrieval}

\paragraph{Retrieval setup and quantitative results.}
We further analyze the learned A2V predictor through audio-to-video retrieval on the VGGSound test set. 

\begin{wraptable}{r}{0.35\textwidth}
\vspace{-1.0em}
\centering
\footnotesize
\setlength{\tabcolsep}{3.0pt}
\renewcommand{\arraystretch}{0.92}
\caption{\textbf{Audio-to-video retrieval on VGGSound for 3B variant.} }
\vspace{-2mm}
\begin{tabular}{lccc}
\toprule
Method & R@1 & R@5  & MedR$\downarrow$ \\
\midrule
\multicolumn{4}{c}{\textit{Qwen2.5-Omni 3B}} \\
\cmidrule(lr){1-4}
orig & 5.2 & 10.0  & 57 \\
ours & \textbf{12.3} & \textbf{36.9} & \textbf{10} \\
\midrule
\multicolumn{4}{c}{\textit{Video-SALMONN2+ 3B}} \\
\cmidrule(lr){1-4}
orig & 1.3 & 4.2 & 101 \\
ours & \textbf{6.5} & \textbf{21.0}& \textbf{21} \\

\bottomrule
\end{tabular}
\label{supp;tab:retrieval_predictor}
\vspace{-1em}
\end{wraptable}
This experiment is used only for predictor analysis and is separate from the downstream pruning benchmarks in the main paper. Because our goal is to evaluate coarse semantic prediction rather than exact instance retrieval, we construct the retrieval candidate set by sampling one video per category.
For each backbone, we compare two audio-side representations: the original audio embedding produced by the backbone audio encoder (\textit{orig}) and the embedding produced by our trained A2V predictor (\textit{ours}). Retrieval is performed by computing cosine similarity between the audio-side representation and the visual embedding of each candidate video. As shown in Table~\ref{supp;tab:retrieval_predictor}, the same trend observed in the 7B models also holds for the 3B variants, where the predictor embedding consistently improves Recall@1 and Recall@5 while substantially reducing the median rank.
Qwen2.5-Omni also shows stronger retrieval performance than Video-SALMONN2+ even with the original audio embedding, indicating a stronger initial cross-modal alignment.

\paragraph{Lightweight predictor design.}
We initially experimented with a larger predictor configuration of 256 learnable queries and 4 cross-attention layers. 

\begin{wraptable}{r}{0.50\textwidth}
\vspace{-1.4em}
\centering
\footnotesize
\caption{\textbf{Audio-to-video retrieval on VGGSound.} Our lightweight predictor matches a larger configuration while reducing inference cost.}
\vspace{-2mm}
\begin{tabular}{lccc}
\toprule
Method & R@1 & R@5 & MedR$\downarrow$ \\
\midrule
\multicolumn{4}{c}{\textit{Qwen2.5-Omni 7B}} \\
\cmidrule(lr){1-4}
Orig                       & 5.8  & 17.8 & 44 \\
Large ($Q{=}256$, $N_{layer}{=}4$) & \textbf{13.9} & \textbf{38.5} & \textbf{10} \\
Ours ($Q{=}128$, $N_{layer}{=}2$)  & 12.9 & 36.9 & 11 \\
\midrule
\multicolumn{4}{c}{\textit{Video-SALMONN2+ 7B}} \\
\cmidrule(lr){1-4}
Orig                       & 2.3 & 8.4  & 68 \\
Large ($Q{=}256$, $N_{layer}{=}4$) & \textbf{8.7} & \textbf{26.5} & \textbf{16} \\
Ours ($Q{=}128$, $N_{layer}{=}2$)  & \textbf{8.7} & 24.0 & 21 \\
\bottomrule
\end{tabular}
\vspace{-1.3em}
\label{supp:light_a2vpred}
\end{wraptable}
However, as shown in Table~\ref{supp:light_a2vpred}, we found that scaling down to 128 queries and 2 cross-attention layers maintains comparable audio-to-video retrieval performance while reducing the predictor's computational footprint. This indicates that the predictor does not require high capacity to capture coarse audio-shared visual semantics, since its supervision is applied only at the global semantic level (mean-pooled representations) rather than at the patch level. We therefore adopt the smaller configuration as our default, consistent with our goal of an inference-time pruning framework that introduces minimal overhead.

\begin{figure*}[t]
    \centering
    \begin{subfigure}[t]{0.98\textwidth}
        \centering
        \includegraphics[width=\textwidth]{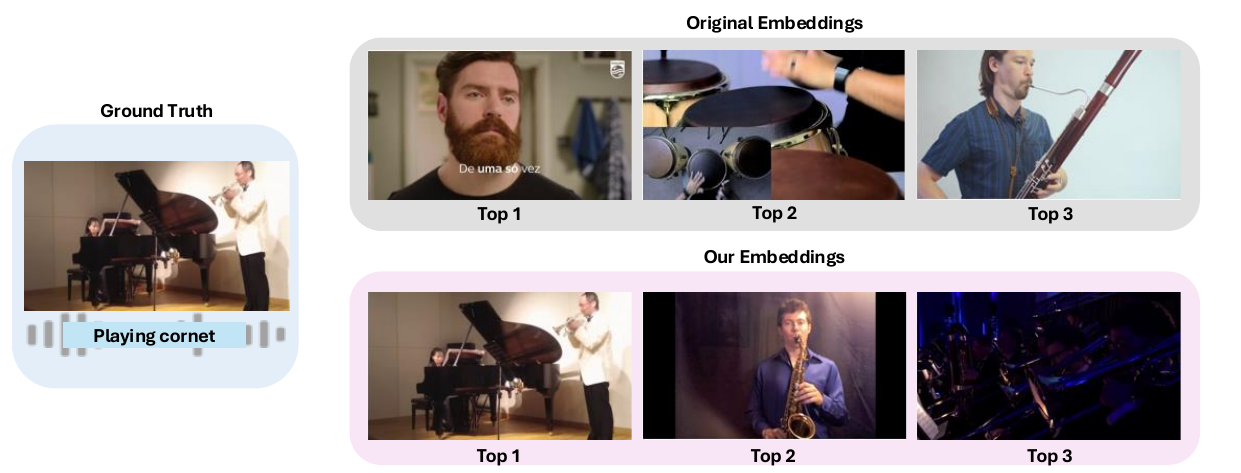}
    \end{subfigure}
    \begin{subfigure}[t]{0.98\textwidth}
        \centering
        \includegraphics[width=\textwidth]{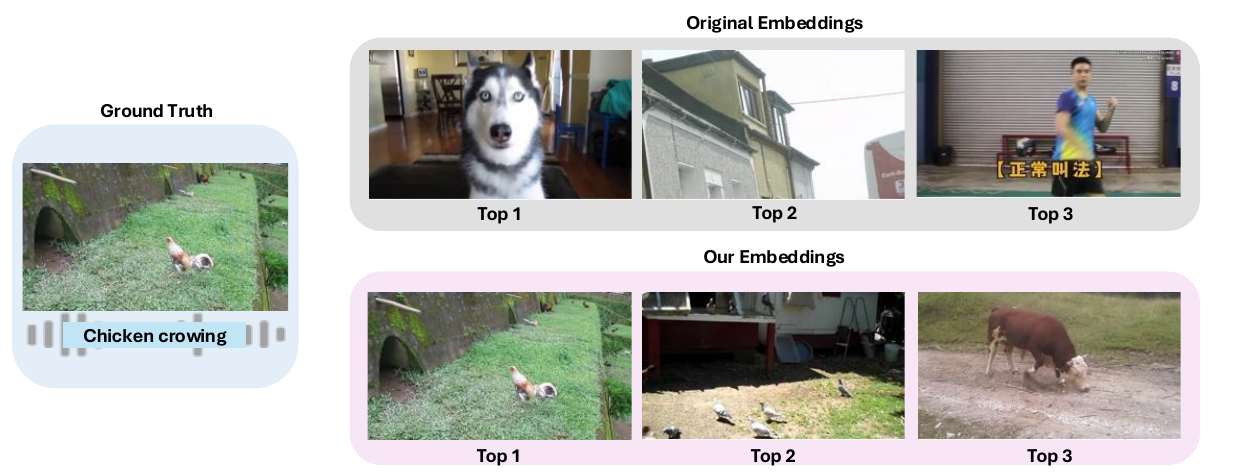}
        
    \end{subfigure}
    \caption{\textbf{Qualitative audio-to-video retrieval results using Qwen2.5-Omni 7B.}}
    \label{supp:fig:ret_qwen}
\end{figure*}

\begin{figure*}[t]
    \centering
    \begin{subfigure}[t]{0.98\textwidth}
        \centering
        \includegraphics[width=\textwidth]{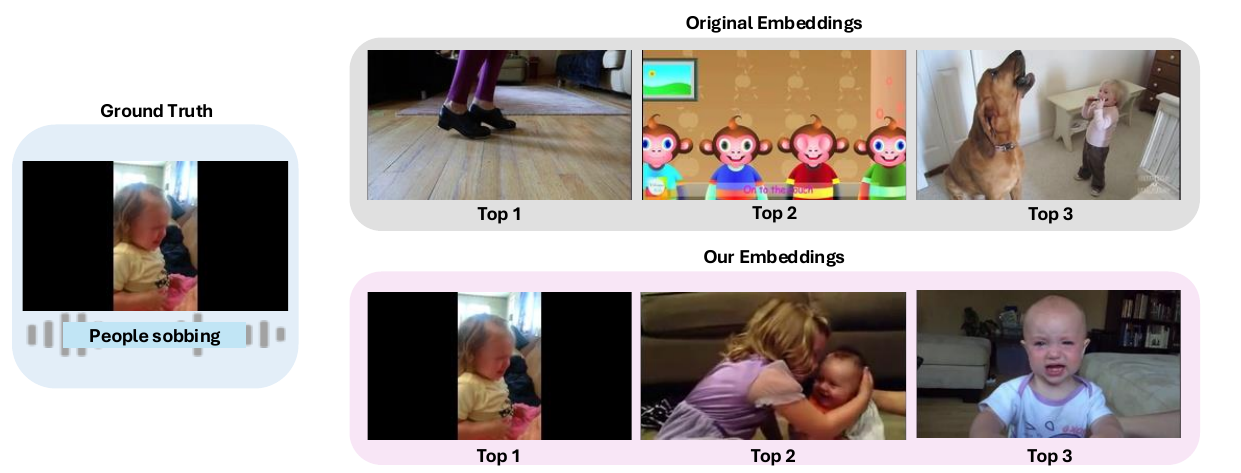}
    \end{subfigure}
    \begin{subfigure}[t]{0.98\textwidth}
        \centering
        \includegraphics[width=\textwidth]{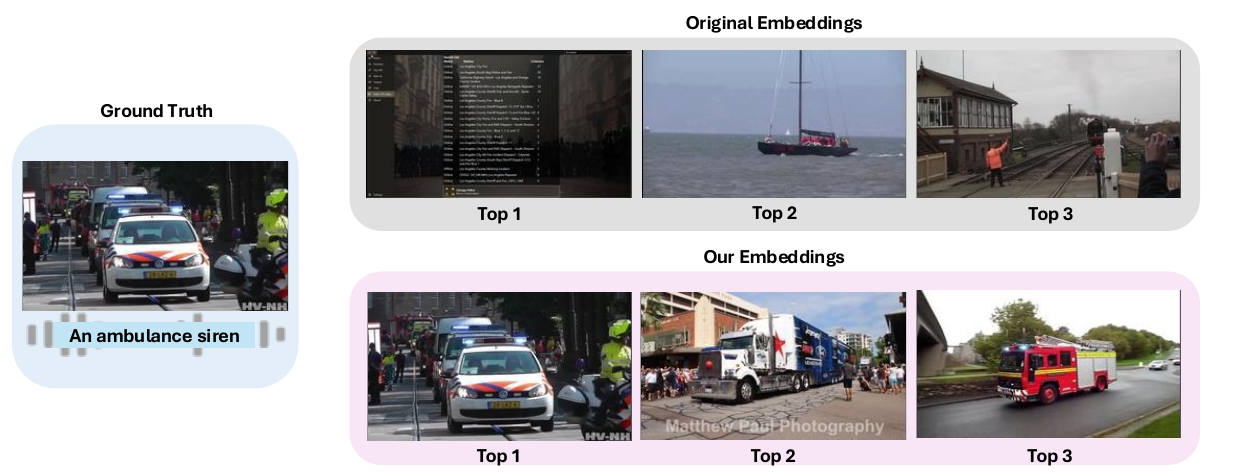}
    \end{subfigure}
    \caption{\textbf{Qualitative audio-to-video retrieval results using Video-SALMONN2+ 7B.}}
    \label{supp:fig:ret_sal}
\end{figure*}

\paragraph{Qualitative results.}
As shown in Figures~\ref{supp:fig:ret_qwen} and~\ref{supp:fig:ret_sal}, we further compare qualitative retrieval examples using the original audio embeddings and the embeddings produced by our A2V predictor. For both Qwen2.5-Omni and Video-SALMONN2+ backbones, the predictor embeddings not only retrieve the ground-truth video more reliably, but also rank semantically similar videos consistently within the top three results.

\clearpage

\clearpage

\section{Analysis on Token Selection Components}
\label{supp:B}
\subsection{Semantic Retention Ratio}

\begin{figure*}[t]
        \centering
        \includegraphics[width=0.6\textwidth]{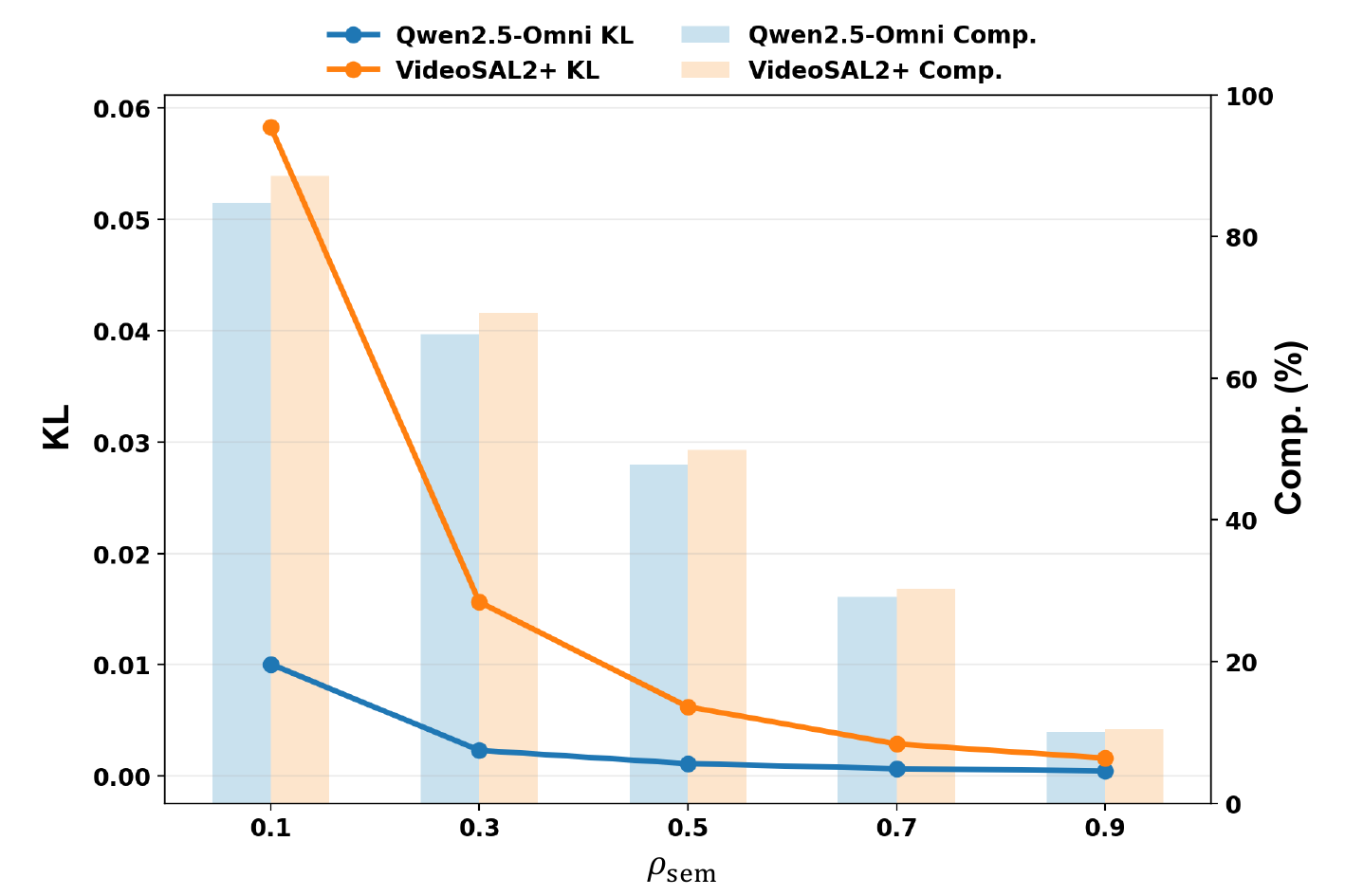}
        \caption{\textbf{Analysis of the semantic retention ratio $\rho_{\mathrm{sem}}$.} Larger $\rho_{\mathrm{sem}}$ values reduce KL divergence to the full-token output distribution by retaining more tokens, but also weaken compression. We choose $\rho_{\mathrm{sem}}=0.5$ for all models as it already achieves low KL divergence while preserving substantial token reduction.}
        \label{fig:app_sem}
\end{figure*}

To select an appropriate semantic retention ratio $\rho_{\mathrm{sem}}$, we analyze how it affects both compression and the deviation from the full-token output distribution. Specifically, we construct an analysis set by randomly sampling 100 examples from the AVQA training set for 10 independent trials. For each value of $\rho_{\mathrm{sem}}$, we retain the bottom-$\rho_{\mathrm{sem}}$ fraction of semantic-similarity tokens and measure both the resulting compression ratio and the KL divergence to the full-token output distribution. As shown in Figure~\ref{fig:app_sem}, increasing $\rho_{\mathrm{sem}}$ reduces KL divergence because more tokens are retained, but the compression benefit correspondingly decreases. We therefore choose $\rho_{\mathrm{sem}}=0.5$, which already achieves low KL divergence while still removing roughly half of the tokens.
\label{supp;sem_ratio}

\subsection{Spatial Detail Preservation}
\label{supp:spa_ratio}
\begin{table}[t]
\centering
\footnotesize
\caption{\textbf{Ablation of spatial detail preservation strategies.} Under the same additional spatial budget, selecting tokens by local spatial variation matches or outperforms random selection and retaining highly audio-aligned tokens.}
\setlength{\tabcolsep}{5pt}
\renewcommand{\arraystretch}{1.12}
\begin{tabular}{llcc}
\toprule
Backbone & Spatial token selection & World. & Daily. \\
\midrule
\multirow{3}{*}{Qwen2.5-Omni 7B}
& Random & \textbf{47.7} & 57.0 \\
& Top-semantic similarity  & 47.3 & 57.1 \\
& Top spatial variation (ours) & \textbf{47.7} & \textbf{57.2} \\
\midrule
\multirow{3}{*}{Video-SALMONN2+ 7B}
& Random & 49.3 & 54.0 \\
& Top-semantic similarity & 48.8 & 54.8 \\
& Top spatial variation (ours) & \textbf{50.6} & \textbf{55.5} \\
\bottomrule
\end{tabular}
\label{tab:spatial_ablation}
\end{table}

\paragraph{Details of grid-wise spatial sampling.}
To encourage broad spatial coverage, we select spatial detail tokens with a grid-wise top-$k$ sampling strategy.
Given an $H \times W$ spatial token map and the spatial retention ratio $\rho_{\mathrm{spa}}$, we set the target number of spatial locations to $N_{\mathrm{spa}}=\lfloor \rho_{\mathrm{spa}} HW \rfloor$.
We then set $g=\lfloor\sqrt{N_{\mathrm{spa}}}\rfloor$ and use strides $\Delta_H=\lfloor H/g \rfloor$ and $\Delta_W=\lfloor W/g \rfloor$ to partition the spatial map into approximate grid cells.
Within each cell, we retain the token with the highest spatial variation score.
Since the cells are generated by stepping over the spatial map, the number of selected locations may slightly differ from $N_{\mathrm{spa}}$ when $H$ or $W$ is not divisible by $g$.
Therefore, $N_{\mathrm{spa}}$ is used as a target budget for determining the grid resolution, rather than as an exact cardinality constraint.
The selected spatial locations are then repeated across frames in the chunk, so $\rho_{\mathrm{spa}}$ controls the fraction of spatial positions retained per frame rather than the total number of spatio-temporal tokens.

\paragraph{Design rationale for the spatial branch.}
For the spatial detail preservation strategy in Sec.~\ref{sec:a2v_token_selection}, we design the spatial branch as a coverage constraint that complements audio-guided semantic pruning.
Under the union-based selection in Eq.~\eqref{eq:framewise_selected_tokens}, low-similarity tokens are already retained by $\mathcal{P}^{(t)}_{\mathrm{sem}}$, so spatial tokens overlapping with $\mathcal{P}^{(t)}_{\mathrm{sem}}$ do not change the final selected set.
As a result, the spatial branch affects the final selection mainly through tokens outside $\mathcal{P}^{(t)}_{\mathrm{sem}}$, i.e., the remaining high-similarity tokens that would otherwise be discarded by semantic pruning.
These tokens may have coarse semantics that are predictable from audio, but can still contain localized visual details, such as color, texture, expression, or pose, that audio alone does not specify.
We therefore apply a simple spatial variation criterion within this remaining token set to retain spatially distributed local details.

Empirically, on WorldSense with Qwen2.5-Omni 7B, the IoU between low-semantic-similarity tokens and spatial-detail tokens is only 12.6\%, indicating that the two branches select largely distinct tokens.
This separation supports our design: semantic pruning retains visual evidence not predicted from audio, while the spatial branch supplements it with localized details from otherwise discarded high-similarity tokens.

To further evaluate the spatial detail preservation strategy, we compare our spatial variation-based selection with two alternatives that retain the same number of tokens: random selection and selection by high semantic similarity to the audio-predicted visual semantics.
As shown in Table~\ref{tab:spatial_ablation}, selecting tokens with high spatial variation matches or outperforms both alternatives on Qwen2.5-Omni and Video-SALMONN2+.
Together, these results indicate that spatial variation is an effective proxy for retaining localized visual details from the remaining audio-explainable tokens, rather than merely selecting tokens with high audio similarity.

\subsection{Implementation Details of Depth-score-based Temporal Merging}
\label{supp:depth_score}
Boundary selection in Sec.~\ref{sec:temporal_merge} uses a fixed threshold on the depth score: a chunk index $t$ is treated as a boundary candidate when $d^{m}_{t} > 0.5$ for either modality $m \in \{v, a\}$.
The depth score in Eq.~\ref{eq:depth_score} is computed only for chunks where both maxima terms are defined, i.e., for $t \in \{2, \ldots, T-2\}$; the first and last adjacent-chunk similarities are excluded from boundary detection by setting their depth scores to zero.
The union of candidates from both modalities partitions the video into temporal segments $\mathcal{S} = \{\mathcal{S}_k\}_{k=1}^{K_s}$.
Within each segment $\mathcal{S}_k$, chunk merging proceeds in a single greedy pass over consecutive chunks using the visual adjacent-chunk similarity $s^{v}_{t} = \mathrm{sim}(\bar{\mathbf{h}}^{v}_{t}, \bar{\mathbf{h}}^{v}_{t-1})$ defined in Sec.~\ref{sec:temporal_merge}. We extend the current group as long as each successive pair satisfies $s^{v}_{t} > \tau_{\mathrm{merge}}$, restarting the group at any pair that falls below the threshold. For example, if chunks $A$, $B$, $C$ in the same segment satisfy $\mathrm{sim}(A, B) > \tau_{\mathrm{merge}}$ and $\mathrm{sim}(B, C) > \tau_{\mathrm{merge}}$, all three are merged into a single representation by averaging their retained token embeddings indexed by $\mathcal{P}^{(k)}_{\mathrm{sel}}$.

\subsection{Hyperparameter Analysis}
\label{supp:hyper_analysis}

\begin{figure*}[t]
    \centering
    \begin{subfigure}[t]{0.98\textwidth}
        \centering
        \includegraphics[width=\textwidth]{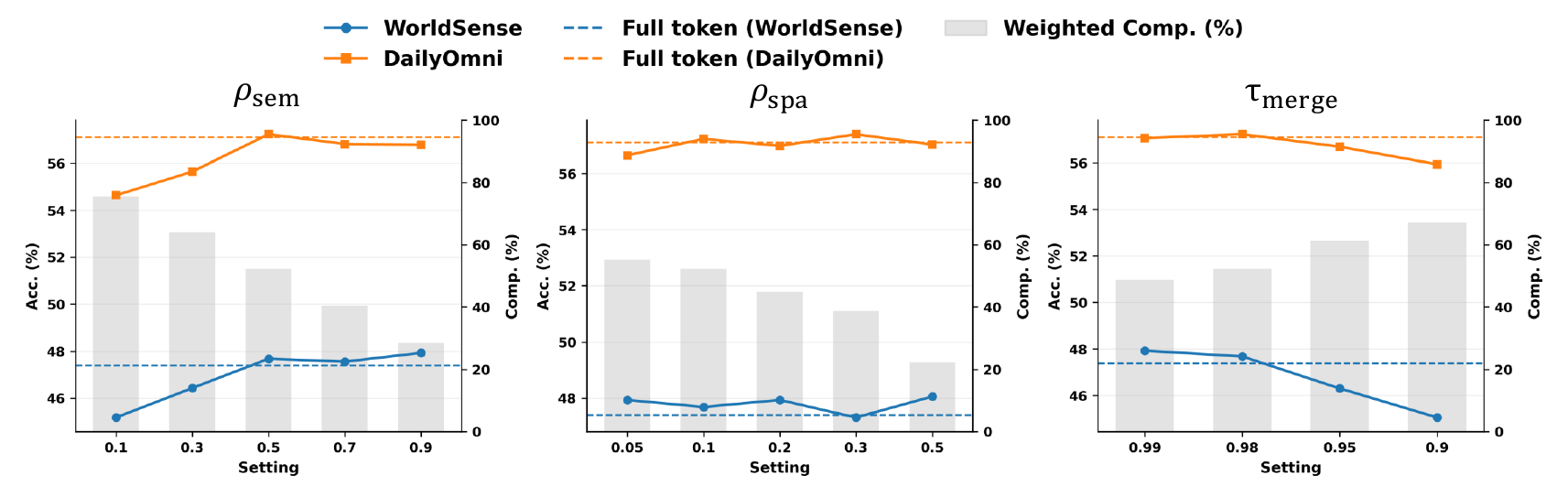}
        \caption{Hyperparameter analysis for Qwen2.5-Omni 7B.}
        \label{fig:app_qwen_hyper}
    \end{subfigure}
    \begin{subfigure}[t]{0.98\textwidth}
        \centering
        \includegraphics[width=\textwidth]{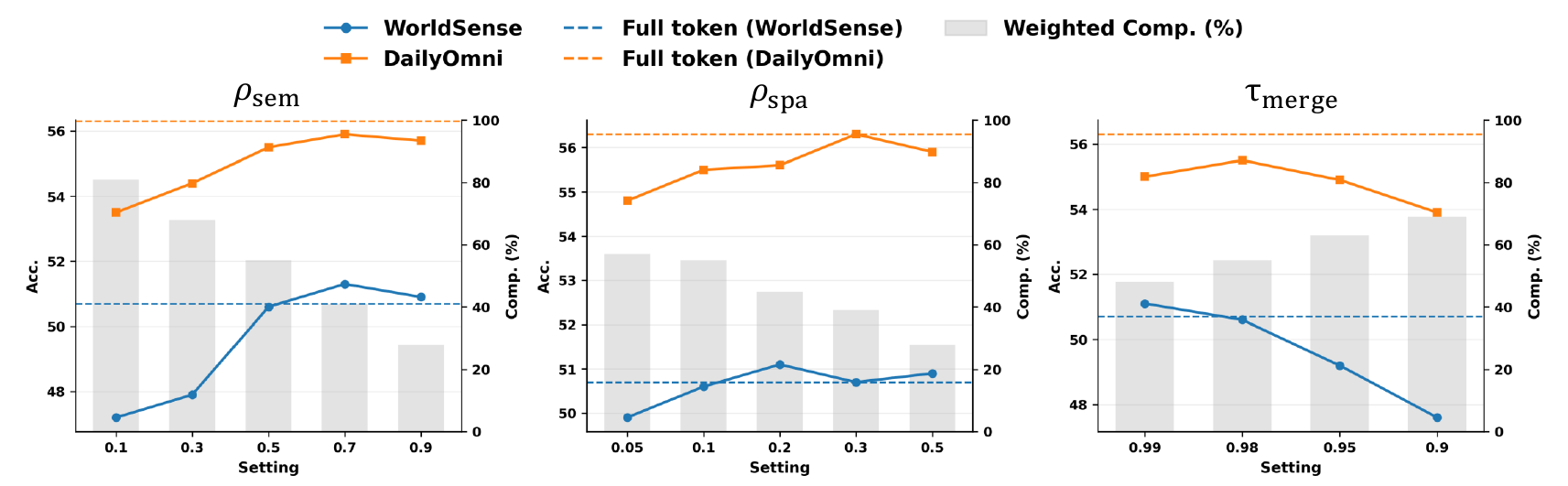}
        \caption{Hyperparameter analysis for video-SALMONN2+ 7B.}
    \end{subfigure}

    \caption{\textbf{Hyperparameter analysis.}}
    \label{fig:app_sal2_hyper}
\end{figure*}

ContextGuard uses three main hyperparameters. First, it retains the bottom-$\rho_{\mathrm{sem}}$ fraction of tokens ranked by semantic similarity to the audio-predicted visual semantics. Second, it preserves the top-$\rho_{\mathrm{spa}}$ fraction of spatial-detail tokens that are less likely to be recoverable from audio alone. Third, within each depth-score-based temporal segment, consecutive chunks are merged when their similarity exceeds the threshold $\tau_{\mathrm{merge}}$. In all main experiments, we fix these hyperparameters to $\rho_{\mathrm{sem}}=0.5$, $\rho_{\mathrm{spa}}=0.1$, and $\tau_{\mathrm{merge}}=0.98$. Figure~\ref{fig:app_sal2_hyper} presents ablations over these choices.

Figure~\ref{fig:app_sal2_hyper} reports hyperparameter ablations on Qwen2.5-Omni 7B and Video-SALMONN2+ 7B using the WorldSense and Daily-Omni benchmarks. Increasing $\rho_{\mathrm{sem}}$ retains more tokens and generally improves fidelity to the full-token model, although downstream accuracy is not strictly monotonic. Our choice of $\rho_{\mathrm{sem}}=0.5$ already matches the full-token baseline on Qwen2.5-Omni while still achieving more than 50\% compression. The spatial retention ratio $\rho_{\mathrm{spa}}$ shows a similar trend: as more spatial detail tokens are preserved, performance generally improves, especially on Video-SALMONN2+. Considering the balance between accuracy and compression, $\rho_{\mathrm{spa}}=0.1$ provides a favorable operating point. Finally, lowering the merge threshold $\tau_{\mathrm{merge}}$ increases compression but can substantially degrade performance. We find that $\tau_{\mathrm{merge}}=0.98$ offers a good balance between temporal compression and downstream accuracy.

Importantly, the same hyperparameter values are fixed across all experiments on Qwen2.5-Omni 7B/3B and Video-SALMONN2+ 7B/3B, as well as all six evaluation benchmarks. This suggests that ContextGuard is reasonably robust to hyperparameter choices and can be used in a practical plug-and-play manner.


\begin{table}[t]
\centering
\footnotesize
\setlength{\tabcolsep}{3.5pt}
\renewcommand{\arraystretch}{1.05}
\caption{\textbf{Category-wise breakdown on Daily-Omni using Qwen2.5-Omni 7B.} ContextGuard shows the largest gains over OmniZip on context-heavy and reasoning-oriented question types, while OmniZip remains stronger only on AV Event Alignment.}
\vspace{0.5em}
\label{tab:dailyomni_breakdown}
\begin{tabular}{lcccccc}
\toprule
\multirow{2}{*}{Method} & AV Event  & Event & \multirow{2}{*}{Reasoning} & \multirow{2}{*}{Inference} & \multirow{2}{*}{Comparative} & Context  \\
& Alignment & Sequence &  &  &  &  understanding \\
\midrule
Full    & \textbf{45.38} & \textbf{50.33} & 73.14          & 75.32          & \textbf{63.36} & 49.22          \\
OmniZip & 43.70          & 49.67          & 73.71          & 76.62          & 61.07          & 48.70          \\
Ours    & 41.18          & 49.67          & \textbf{74.86} & \textbf{79.22} & \textbf{63.36} & \textbf{51.30} \\
\midrule
$\Delta_{\text{Ours-OmniZip}}$ & -2.52          & 0.00           & \textbf{+1.15}          & \textbf{+2.60}          & \textbf{+2.29}          & \textbf{+2.60}          \\
\bottomrule
\end{tabular}
\end{table}

\section{Additional Experimental Details and Analysis}

\subsection{Category-wise Breakdown on Daily-Omni}
\label{supp:dailyomni_breakdown}

To further analyze where ContextGuard is most effective, we provide a category-wise breakdown on Daily-Omni using Qwen2.5-Omni 7B. As shown in Table~\ref{tab:dailyomni_breakdown}, ContextGuard achieves the largest gains over OmniZip on context-heavy and reasoning-oriented question types, including \textit{Reasoning} and \textit{Inference} for cross-modal reasoning, \textit{Context understanding} for scene-grounded interpretation, and \textit{Comparative} for comparing audio-visual cues. In contrast, OmniZip is stronger only on \textit{AV Event Alignment}, which is consistent with its alignment-focused design. This pattern supports our claim that ContextGuard is particularly beneficial when answering requires broader audio-visual context.

\subsection{Additional Experimental Details}
For long-video benchmarks, we restrict evaluation to QA pairs whose source videos are shorter than 1 minute, as full-token evaluation on the complete benchmark exceeds our available memory.
The resulting subset contains 799 QA pairs from 403 videos for WorldSense, 231 QA pairs from 77 videos for Video-MME, and 166 QA pairs from 96 videos for OmniVideoBench.
Although the clips are shorter than one minute, each sample still typically contains several thousand high-resolution visual tokens, making the benchmark suitable for validating token pruning methods.

FastV was originally designed for vision-language models where pruning is applied over visual tokens.
In our AV setting, we apply the same importance scoring mechanism to the expanded audio-video token sequence without modifying the original FastV pruning criterion.

We do not include EchoingPixels and OmniSIFT in the main quantitative comparison because, to the best of our knowledge, their official implementations or checkpoints are not publicly available at the time of submission.
We therefore focus on reproducible baselines with available implementations, including OmniZip, the most closely related recent inference-time omnimodal pruning method, and FastV, a representative video-LLM pruning baseline.

\subsection{Qualitative Results on Downstream QA}
\label{supp:downstream_qual}

\begin{figure}[t]
    \centering
        \centering
        \includegraphics[width=\textwidth]{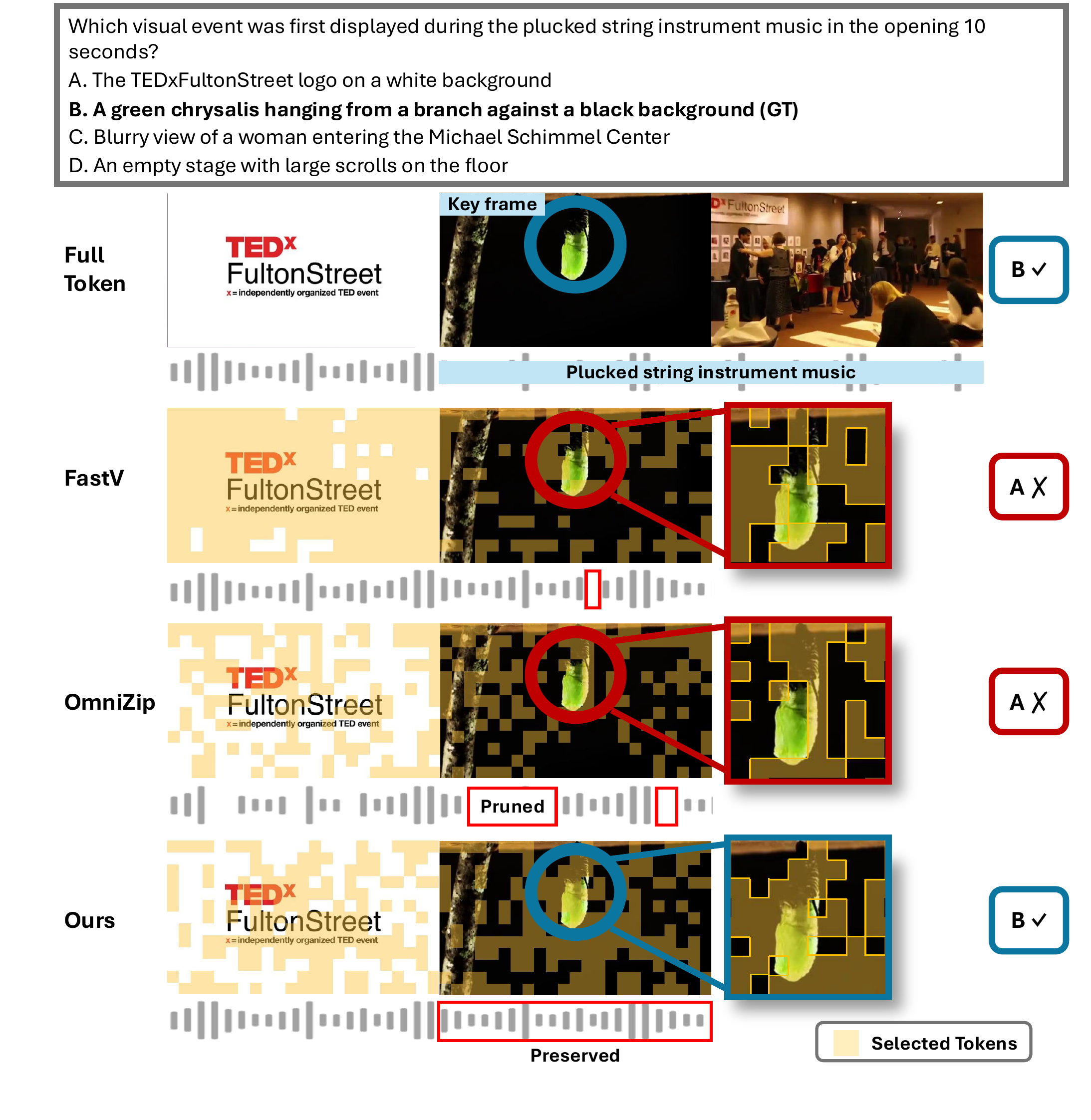}
    \caption{\textbf{Additional qualitative results on downstream QA using Qwen2.5-Omni 7B.} ContextGuard preserves broad AV context and recovers the correct answer.}
    \label{fig:supp_qual1}
\end{figure}

\begin{figure}[t]
    \centering
        \centering
        \includegraphics[width=\textwidth]{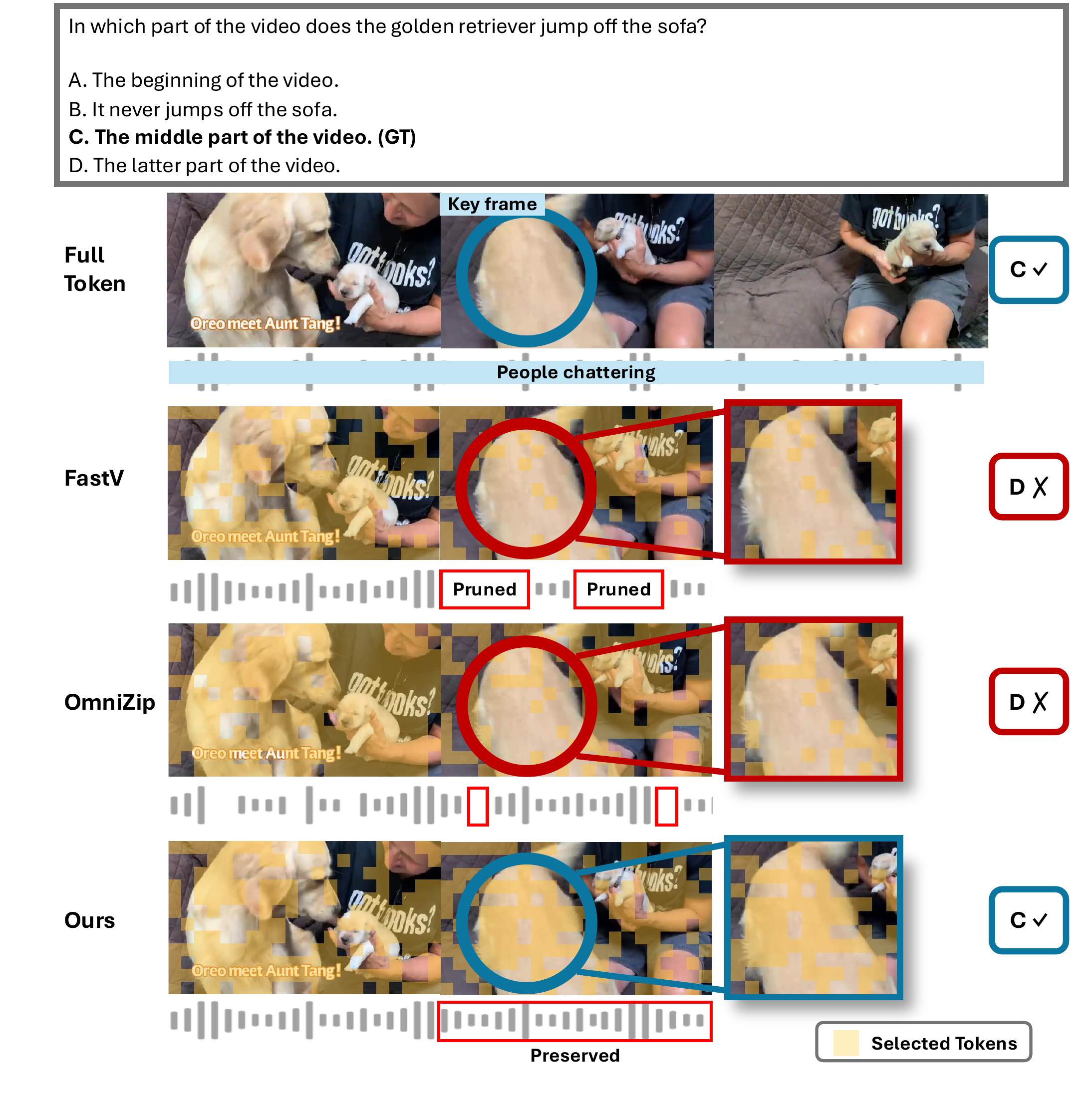}
    \caption{\textbf{Additional qualitative results on downstream QA using Video-SALMONN2+ 7B.} ContextGuard preserves non-audio-aligned visual events, maintains broad AV context, and recovers the correct answer.}
    \label{fig:supp_qual2}
\end{figure}

\begin{figure}[t]
    \centering
        \centering
        \includegraphics[width=\textwidth]{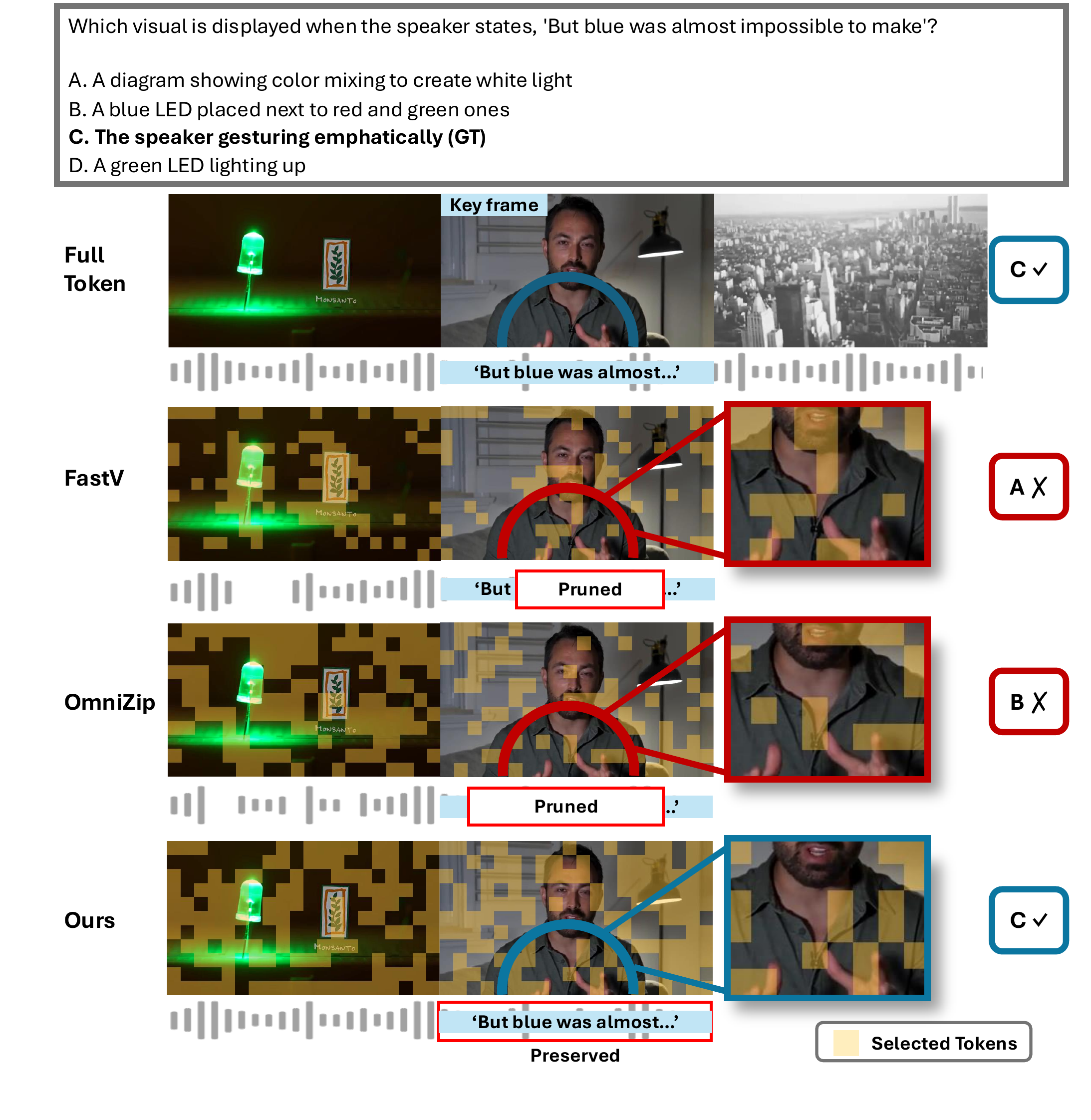}
    \caption{\textbf{Additional qualitative results on downstream QA using Video-SALMONN2+ 7B.} ContextGuard preserves the full speech cue and broad AV context, and recovers the correct answer.}
    \label{fig:supp_qual3}
\end{figure}

\paragraph{Additional qualitative results.}
We further analyze qualitative downstream QA examples using both Qwen2.5-Omni 7B and Video-SALMONN2+ 7B, as shown in Figures~\ref{fig:supp_qual1}--\ref{fig:supp_qual3}.

In Figure~\ref{fig:supp_qual1}, the question requires identifying the visual event that occurs when the audio contains ``plucked string instrument music.'' Although FastV preserves the audio cue better than OmniZip, both methods fail to retain the green chrysalis and therefore predict the wrong answer. In contrast, ContextGuard preserves both the relevant audio cue and the non-audio-aligned visual evidence, thereby maintaining broad context and leading to the correct answer.

The example in Figure~\ref{fig:supp_qual2} requires temporally localizing the event where the dog jumps off the sofa. 
Because the audio mainly contains people chattering and lacks dog-related cues, FastV and OmniZip fail to consistently preserve the dog, likely favoring visually salient or audio-anchored regions instead. By contrast, ContextGuard better preserves the dog and the broader AV context, thereby recovering the correct answer.

Figure~\ref{fig:supp_qual3} requires both accurate speech localization for the phrase ``But blue was almost ...'' and recognition of the temporally aligned visual content. FastV and OmniZip prune the relevant speech segment and preserve only limited gesture-related patches, which leads to failure. In contrast, ContextGuard retains the necessary speech cue together with sufficient gesture-related visual evidence and produces the correct answer. This example further illustrates that audio tokens provide highly compact yet critical information, and highlights the advantage of ContextGuard, which preserves the full audio stream while still achieving high compression ratios relative to prior pruning baselines.

\paragraph{Failure-case analysis.}
To further analyze the limitations of ContextGuard, we examine failure cases on the WorldSense benchmark using both backbones.

\begin{figure}[t]
    \centering
        \centering
        \includegraphics[width=\textwidth]{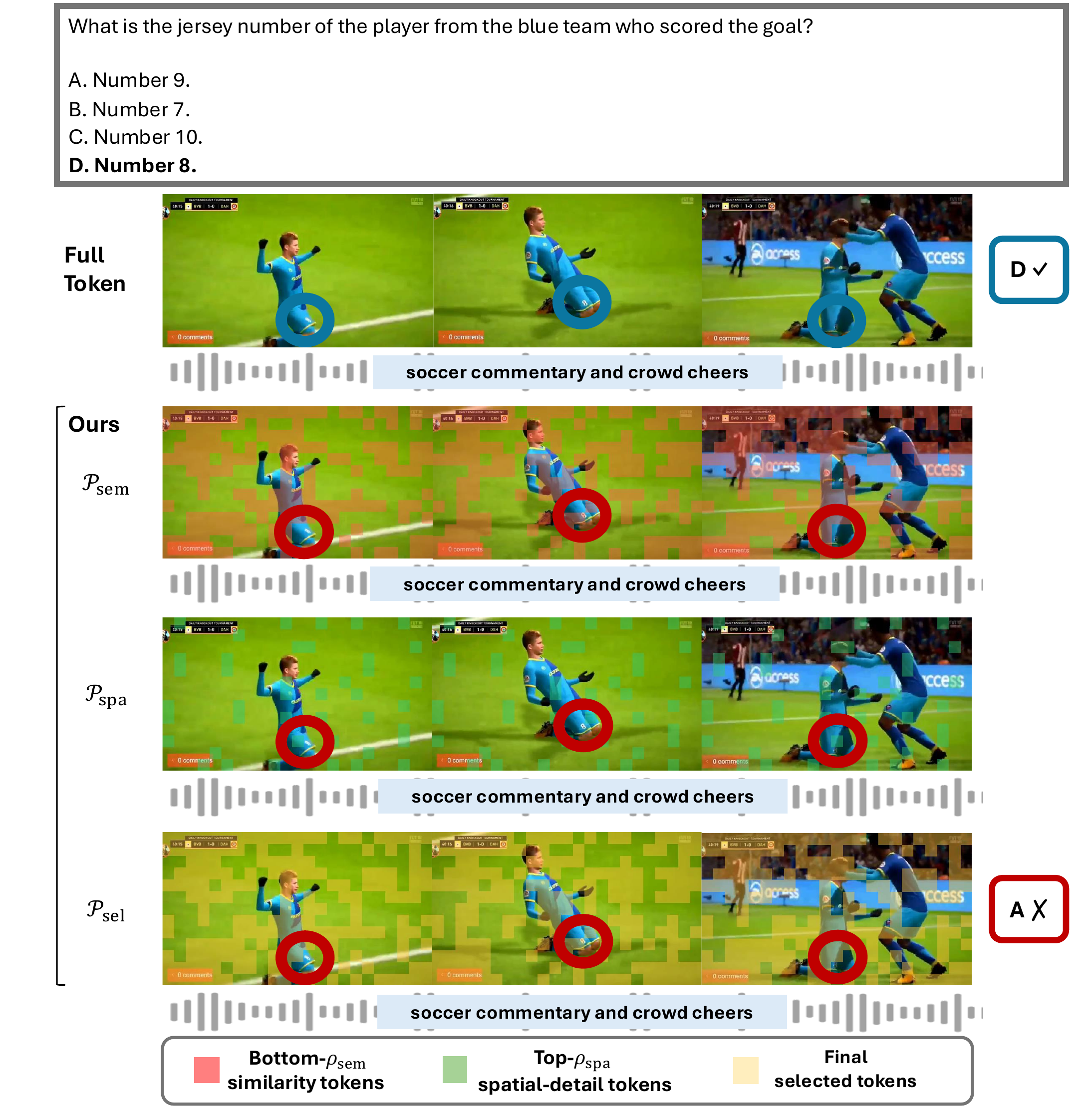}
   \caption{\textbf{Failure case on downstream QA using Qwen2.5-Omni 7B.}
ContextGuard misses a subtle fine-grained detail, the player's jersey number, leading to an incorrect answer.}
    \label{fig:supp_fail_1}
\end{figure}

As shown in Figure~\ref{fig:supp_fail_1}, the question requires a fine-grained visual detail, namely the player's jersey number. 
ContextGuard does not consistently preserve this detail across video chunks, resulting in unstable predictions and incorrect answers. 
This suggests that retaining a fixed top-$10\%$ of spatial-detail tokens may be insufficient to preserve all fine-grained visual evidence in some chunks.

\begin{figure}[t]
    \centering
        \centering
        \includegraphics[width=\textwidth]{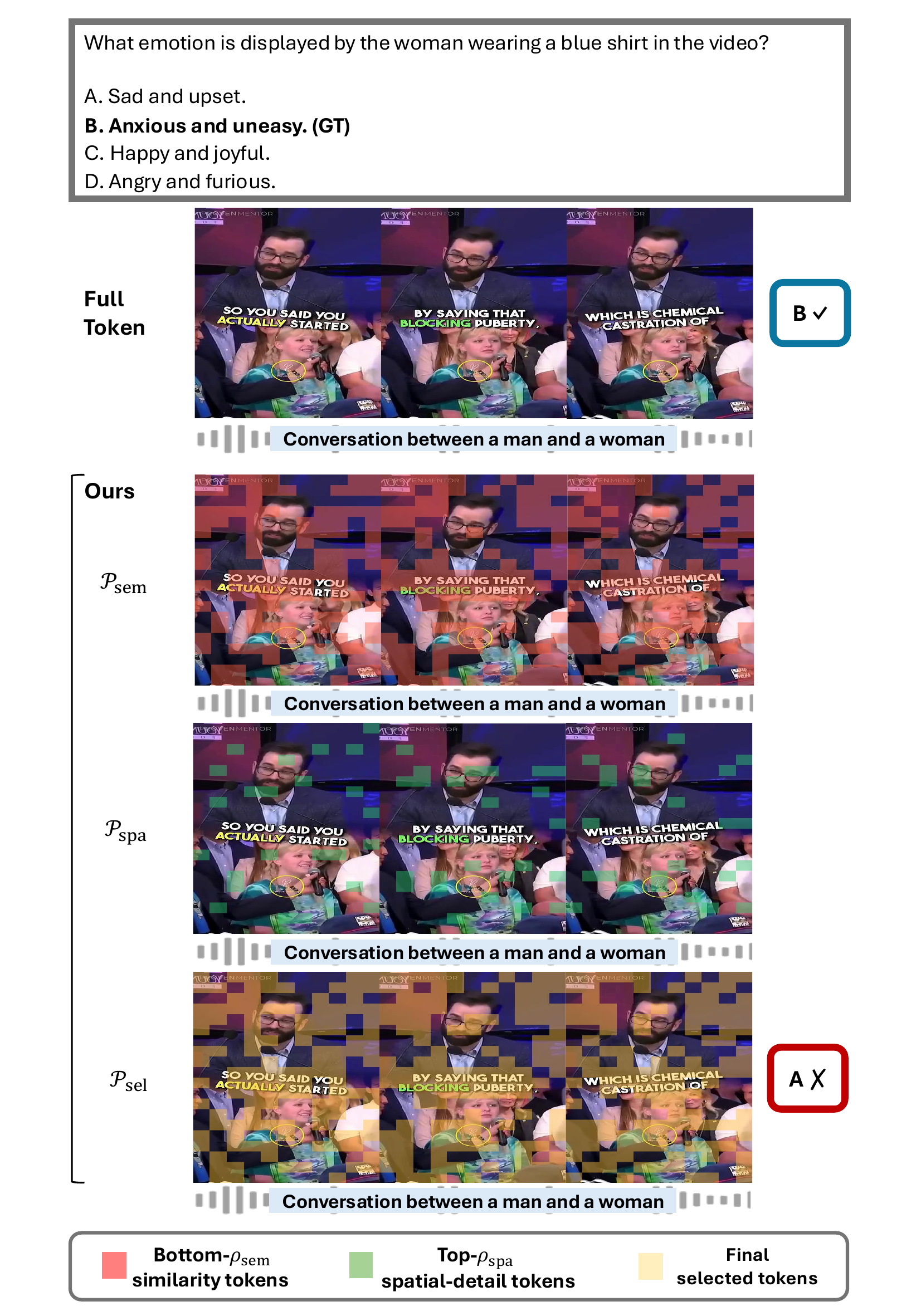}
    \caption{\textbf{Failure case on downstream QA using Video-SALMONN2+ 7B.}
ContextGuard preserves evidence not recoverable from audio, such as OCR text, but fails to consistently retain fine-grained temporal visual cues, such as facial expressions, needed for the correct answer.}
    \label{fig:supp_fail_2}
\end{figure}

As shown in Figure~\ref{fig:supp_fail_2}, the question requires understanding both the conversation and the woman's facial expression. 
Although ContextGuard preserves relevant OCR text and prunes the audio-dominant male speaker, it sometimes fails to retain sufficient fine-grained facial-expression cues for the correct answer.

These cases suggest that while ContextGuard effectively reduces visual tokens with audio-redundant coarse semantics, a fixed spatial retention budget may not always capture subtle fine-grained evidence.
In future work, we plan to address this limitation by adaptively selecting $\rho_{\mathrm{spa}}$ according to the input context.

\clearpage

\subsection{Qualitative Analysis of Non-Audio-Aligned Semantic Selection}
\label{supp:qual_semantic}
We further examine whether the bottom-$\rho_{\mathrm{sem}}$ low-similarity semantic tokens selected by ContextGuard correspond to visual regions that are not predictable from audio.

\begin{figure*}[t]
    \centering
    \begin{subfigure}[t]{0.75\textwidth}
        \centering
        \includegraphics[width=\textwidth]{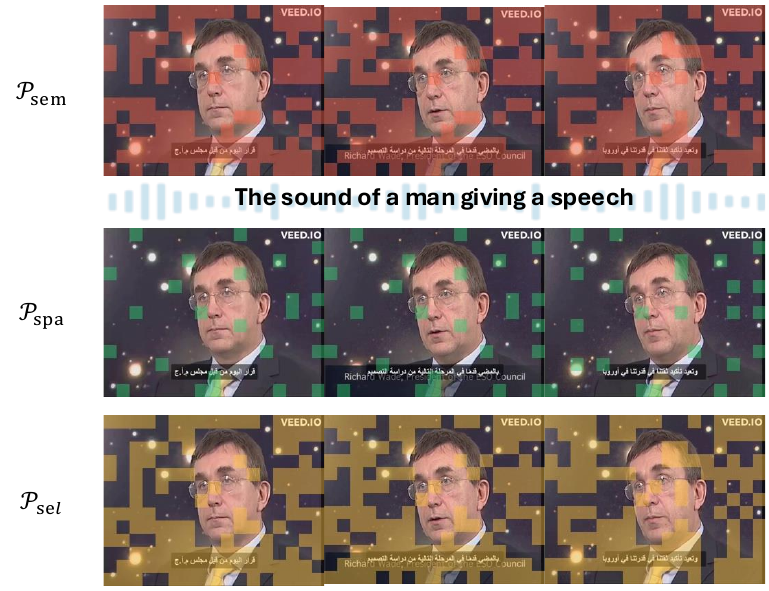}
        \caption{A speaking-person scene with non-audio-aligned text and visual details.}
        \label{supp:qual_sem_qwen_a}
    \end{subfigure}
    \begin{subfigure}[t]{0.7\textwidth}
        \centering
        \includegraphics[width=\textwidth]{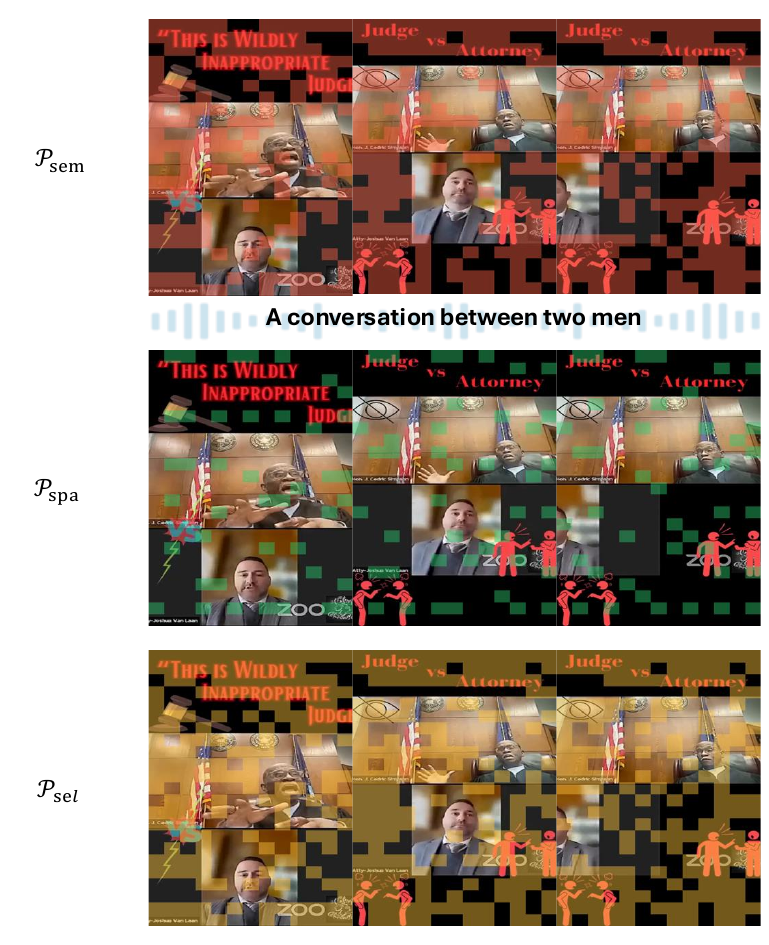}
        \caption{A two-person conversation scene with non-audio-aligned textual evidence.}
        \label{supp:qual_sem_qwen_b}
    \end{subfigure}
    \caption{\textbf{Qualitative analysis of non-audio-aligned semantic selection using Qwen2.5-Omni 7B.} 
    ContextGuard preserves non-audio-aligned semantic regions such as scene text, while the spatial-detail branch further helps retain localized visual details.}
    \label{fig:app_qwen_sem}
    
    \vspace{-3mm}
\end{figure*}

As shown in Figure~\ref{supp:qual_sem_qwen_a}, the video contains a speaking person, whose presence is largely predictable from the audio stream.
ContextGuard therefore selects low-similarity semantic tokens from regions beyond the speaker, consistently preserving these non-audio-aligned parts over time in $\mathcal{P}_{\mathrm{sem}}$.
Notably, ContextGuard also preserves text appearing in the video, which cannot be inferred from audio alone.
The spatial-detail preservation branch further supplements these semantic tokens with localized visual details, such as the speaker's hair color.
Together, these examples show that our pruning strategy can reduce tokens while maintaining broad video content.

Figure~\ref{supp:qual_sem_qwen_b} shows a similar pattern.
Although the video is dominated by a two-person conversation, ContextGuard preserves non-audio-aligned visual evidence such as scene text, for example ``This is Wildly inappropriate judge.''
This again suggests that the semantic selection branch favors visual regions not predicted from audio, rather than only strongly audio-aligned regions.

\begin{figure*}[t]
    \centering
    \begin{subfigure}[t]{0.75\textwidth}
        \centering
        \includegraphics[width=\textwidth]{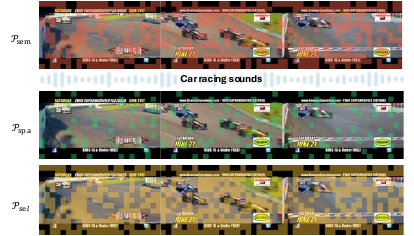}
        \caption{A racing scene with non-audio-aligned visual context.}
        \label{supp:qual_sem_sal2_a}
    \end{subfigure}
    \begin{subfigure}[t]{0.7\textwidth}
        \centering
        \includegraphics[width=\textwidth]{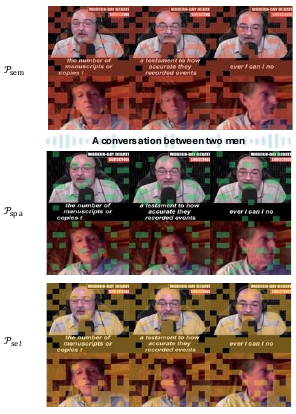}
        \caption{A two-person conversation scene with non-audio-aligned textual evidence.}
        \label{supp:qual_sem_sal2_b}
    \end{subfigure}
    \caption{\textbf{Qualitative analysis of non-audio-aligned semantic selection using Video-SALMONN2+ 7B.} Similar to Qwen2.5-Omni, ContextGuard preserves non-audio-aligned semantic regions while avoiding over-retention of strongly audio-aligned content.}
    \label{fig:app_sal2_sem}
\end{figure*}

Video-SALMONN2+ exhibits the same behavior in Figures~\ref{supp:qual_sem_sal2_a} and~\ref{supp:qual_sem_sal2_b}.
ContextGuard consistently preserves non-audio-aligned regions such as scene text, while avoiding over-retention of strongly audio-aligned regions such as the car in a racing scene or the speaking persons in a two-person conversation.

Together with the category-wise breakdown in App.~\ref{supp:dailyomni_breakdown}, these examples show that ContextGuard preserves non-audio-aligned evidence that supports broad AV understanding.

\section{Extensions and Discussion}

\subsection{Online-Friendly Variant}
\label{supp:online_variant}

Our main offline method performs depth-score-based temporal segmentation and then merges temporally redundant chunks within each segment.
Because depth scores require access to the full sequence, this design is not directly applicable to causal or online-friendly settings.

To explore whether the same pruning principle can be adapted to such settings, we additionally evaluate a simple online-friendly variant in Table~\ref{tab:online_variant}.
Instead of using depth-score-based segmentation and segment-level merging, this variant relies only on local chunk-to-chunk similarity.
Specifically, we compare each chunk with its immediate predecessor and discard the previous chunk if their similarity exceeds $0.99$.
This variant operates entirely on the input video sequence before LLM inference, with chunk removal performed during input compression.
Despite its simplified design, the variant shows comparable performance under similar compression, suggesting that the proposed principle may extend to more online-friendly settings.




\subsection{Limitations}
\label{supp:limitations}

ContextGuard adds a lightweight pre-decoding stage with an A2V predictor and depth-score-based temporal merging, introducing some prefill overhead over simpler inference-time pruning baselines, although this is largely offset by the reduced visual token budget. It also preserves broad AV context rather than optimizing a task-specific subset for each fine-grained query, so a fixed spatial retention budget may still miss subtle localized evidence such as jersey numbers or facial expressions. 
Improving predictor efficiency and adapting the spatial retention budget remain promising future directions.

\subsection{Broader Impacts}
\label{supp:broader_impacts}

ContextGuard improves the efficiency of Omni-LLMs by reducing redundant visual tokens, potentially lowering memory usage and inference costs for audio-visual applications.
While this may make multimodal systems more accessible and practical to deploy, it could also lower the barrier to misuse in privacy-sensitive settings, such as large-scale video analysis or surveillance.
We encourage responsible deployment with appropriate data governance, privacy safeguards, and application-specific safety checks.


\end{document}